\documentclass[11pt,twoside,twocolumn,a4paper]{article}

\usepackage{cvww}
\usepackage{times}
\usepackage{epsfig}
\usepackage{graphicx}
\usepackage{amsmath}
\usepackage{amssymb}
\usepackage{subfig}
\usepackage{svg}
\usepackage{booktabs}

\usepackage[pagebackref=true,breaklinks=true,bookmarks=false]{hyperref}

\cvwwfinalcopy 

\def\cvwwPaperID{5} 

\ifcvwwfinal\fi
\begin{document}
\title{Situation-Aware Pedestrian Trajectory Prediction with Spatio-Temporal Attention Model}

\author{Sirin Haddad\and Meiqing Wu\and He Wei\and Siew Kei Lam\\
Nanyang Technological University (NTU)\\
50 Nanyang Ave, Singapore\\
\tt\small{\{siri0005,wei005\}@e.ntu.edu.sg}
\tt\small{\{meiqingwu,assklam\}@ntu.edu.sg}
}

\maketitle
\ifcvwwfinal\thispagestyle{fancy}\fi



\maketitle
\setcounter{page}{1}
\pagenumbering{arabic}
\begin{abstract}

\textit{Pedestrian trajectory prediction is essential for collision avoidance in autonomous driving and robot navigation. However, predicting a pedestrian's trajectory in crowded environments is non-trivial as it is influenced by other pedestrians' motion and static structures that are present in the scene. Such human-human and human-space interactions lead to non-linearities in the trajectories. In this paper, we present a new spatio-temporal graph based Long Short-Term Memory (LSTM) network for predicting pedestrian trajectory in crowded environments, which takes into account the interaction with static (physical objects) and dynamic (other pedestrians) elements in the scene. 
Our results are based on two widely-used datasets to demonstrate that the proposed method outperforms the state-of-the-art approaches in human trajectory prediction. In particular, our method leads to a reduction in Average Displacement Error (ADE) and Final Displacement Error (FDE) of up to 55\% and 61\% respectively over state-of-the-art approaches.}

\end{abstract}


\section{Introduction}

\begin{figure} 
    \centering
    \includegraphics[width=8cm, height=5cm]{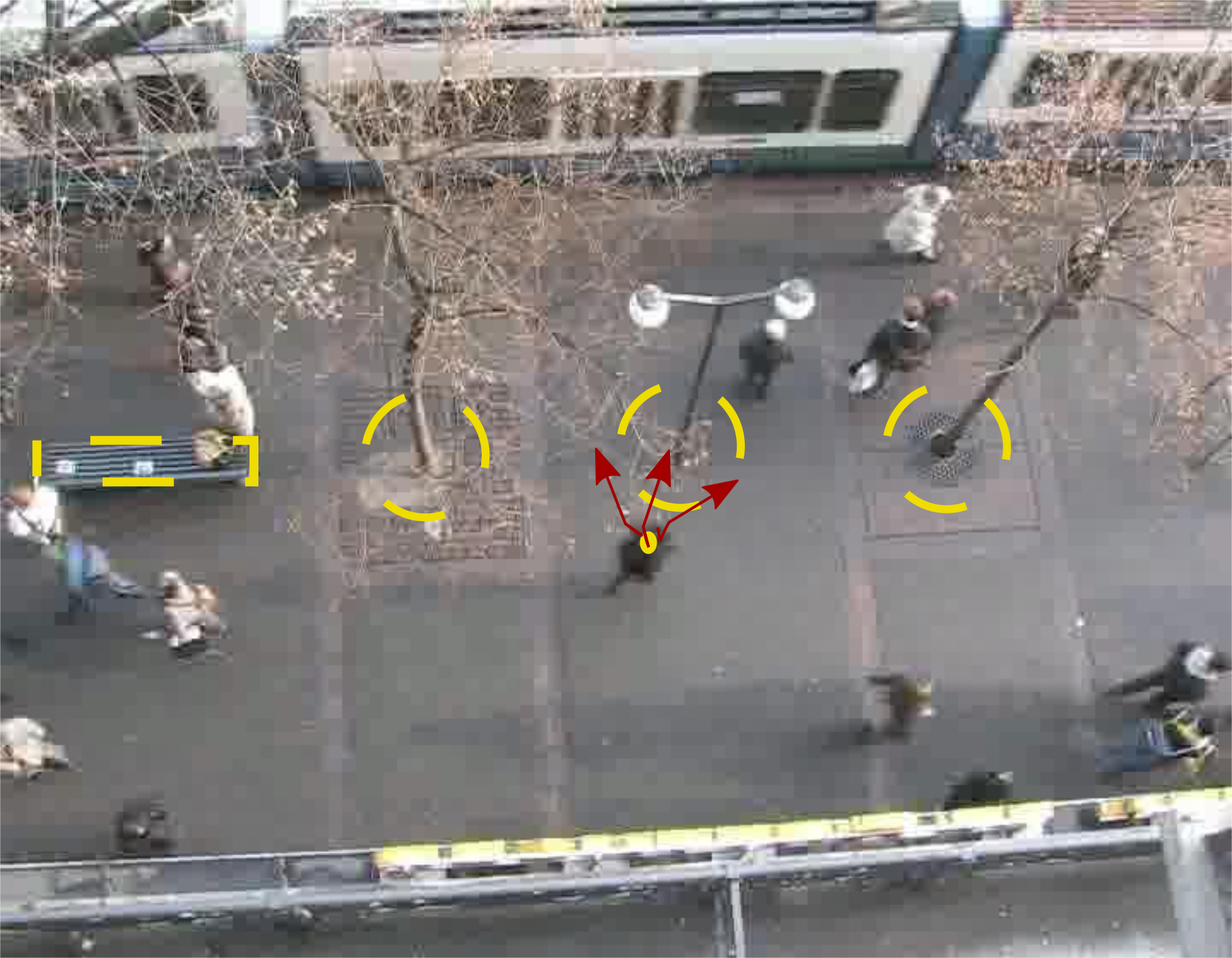}
    \caption{Estimating pedestrian trajectory given the surrounding environment physical structure in a real-life scenario. In the figure above, a lamp post lies in the direction of traversal of the pedestrian of interest. Thus it is essential to capture the existence of static obstacle and understand how they will navigate around it.}
    \label{fig:obstacle_scenario}
\end{figure}

The provision to estimate future trajectories of pedestrians and predicting the possibility of collisions can prevent accidents in autonomous driving and robot navigation. However, pedestrian trajectory prediction in crowded environments is a challenging task as human navigation decisions are influenced by their interactions with other traffic participants and the static physical objects. In particular, humans navigate in a situation-aware manner by avoiding collisions with static objects and other pedestrians in the space surrounding them, based on common social rules. As such, prediction models must take into account the interactions of both static and dynamic elements in the environment in order to accurately predict the pedestrians' motion paths. Figure \ref{fig:obstacle_scenario} shows a real scenario that requires awareness of the lamp post presence in order to make realistic prediction about the pedestrian trajectory who will avoid walking into paths leading to the obstacles area.


Previous works that addressed human motion prediction focused on modeling human-human and human-space interaction separately.
\cite{kretzschmar2014learning,koppula2016anticipating,bartoli2017context, ellis2009modelling,kim2011gaussian} account for scene static configuration such as obstacles and scene structures for improving human trajectory predictions in the presence of dynamic objects. However, these works mainly target constrained environments with low crowd density.

Recently, the work in \cite{robicquet2016learning} presented a deep convolutional network that models the impact of scene static elements on the pedestrian motion. However, they relied on complex tools comprising convolution layers and multiple feature maps for modeling knowledge about the scene. Recurrent neural networks in \cite{vemula2018social,alahi2016social,gupta2018social} tackled pedestrian trajectory prediction on challenging datasets of outdoor scenes \cite{sadeghiankosaraju2018trajnet}. Nevertheless, these approaches only modeled the social interaction among pedestrians without taking into account the surrounding static context.
Social Attention \cite{vemula2018social} encapsulated the social interactions along the spatial and temporal domains by adopting spatio-temporal graph architecture. Their model considered the social interaction as a global event occurring between each and every pedestrian using their velocity to state their influence on each other.
In contrast, Social LSTM \cite{alahi2016social} only accounted for the influences within a fixed-size local neighborhood.

In our work, we propose an enhancement to the models in \cite{vemula2018social,jain2016structural} and improve the modeling of multiple trajectories correlations over space-time dimensions using the 2D locations of the the static and dynamic elements. 
In particular, the proposed model overcomes the limitation of Social LSTM \cite{alahi2016social} which only accounted for the influence of other pedestrians within a local neighborhood, while at the same time being cognizant of the static obstacles at close proximity. This concern was not present in Social Attention \cite{vemula2018social}. Our intuition for this model is that while a pedestrian's trajectory can be affected by the dynamic motions of other pedestrians at a distance, the decision to avoid static objects is usually made when the pedestrian is close to the object. Thus, we manage to reduce the graph complexity and achieve more stable predictions by dynamically incorporating the static elements in the graph structure only when they potentially pose an impact on pedestrian trajectory. 

Our main contributions are as follows: $\left(1\right)$ we present a spatio-temporal graph that explicitly captures the global interaction of all the pedestrians in the scene and the local interaction with the static objects, and  $\left(2\right)$ we propose a new spatio-temporal attention mechanism for each pedestrian trajectory. This mechanism takes into account the local interaction among pedestrians and objects. Our spatio-temporal mechanism is inspired by the work of \cite{velickovic2017graph} which casts the attention methods \cite{vaswani2017attention} for sequence learning tasks on graphs.
Experimental results on two widely-used datasets demonstrate that our method achieves significant quantitative and qualitative improvements over state-of-the-art methods for pedestrian trajectory prediction. 


\section{Related Work}
\label{section:relatedWork}
In this section, we present a summary of research on pedestrians trajectory prediction. 
The literature branches into two main trends regarding context inclusion: local context and global context. 
Additionally, the existing works unfolds into two other branches in terms of distinguishing multiple objects influence: attention-based and uniformly-based approaches. 

\textbf{Local context Versus Global context.}
It is obvious from the previous introduction that the modern trajectory prediction approaches \cite{van2018relational,battaglia2016interaction,alahi2016social,bartoli2017context,fernando2018soft+,helbing1995social} resorted to a limited spatial extent of the surrounding context as they observed the interactions occurring within short distance from the pedestrian included, while \cite{vemula2018social,gupta2018social} were globally-based as they considered all the pedestrians in the scene even those who are far away from each other.

According to local context methods, observing the interaction for a short duration once pedestrians are close enough to each other, gives limited understanding of the social interaction. While including the social interactions on a global scene scale, enables the model to better understand how the interaction evolves between a pair of pedestrians based on the velocity effect that the model inherently grasps upon capturing the change in the spatial distances along time. 

\textbf{Attention-based Versus Uniformly-based approaches.}
Pedestrians navigating in urban environments influence each other and very often are influenced by the obstacles around them, thus it is essential for predicting multiple pedestrians trajectories to recognize the importance of various sources impact on a pedestrian and pay attention to the more influential ones. Applying attention in sequence learning tasks has proved its effectiveness in the overall algorithm performance and in pedestrian trajectory prediction methods it helped drawing more plausible trajectories.

The variational encoder-decoder methods, such as, Social GAN \cite{gupta2018social} took the global neighborhood around pedestrian but it evaluates all pedestrians in a uniform manner, by assigning equal importance values to them. Existing RNN approaches \cite{vemula2018social} applied soft attention mechanism to assign different importance weights to multiple pedestrians based on their velocities. While \cite{fernando2018soft+} applied hard attention to assign weights based on pedestrians distance, they also introduced additional soft attention to evaluate the interaction salience in a scene region. So, their trajectory prediction drew conclusions about which region was more likely for a pedestrian to navigate through. 
In our work, we are rather interested in microscopic prediction of the interaction between pedestrians and a specific fixed obstacle, hence, we use the soft attention mechanism \cite{velickovic2017graph} to evaluate the social interactions only.

\textbf{Graph-Structured Networks.}
Real-life applications generate complicated forms of information in which they are best represented through graph structures compared to other rigid hierarchical and end-to-end organizations.
Variational Encoder-Decoder methods\cite{gupta2018social,varshneya2017human,lee2017desire}, have the advantage of generating a variety of results, however, they are not capable of providing a factorized and explicit high-level representation of the environment components.
Graph Neural Network \cite{scarselli2009graph} advanced the application of graph-structured data in neural networks in environments that naturally contain highly interrelated behaviors, such as: social media, molecular biology, etc. Outdoor pedestrians navigation typically induces a spatio-temporal nature due to alterations that happen in pedestrian motion trajectory and the complex interactions with different objects. 
Therefore, modeling a rich interactive context requires a scalable graph-based structuring of the elements and factorize their relationships in a principled way.
Neural relational networks \cite{van2018relational,battaglia2016interaction}, attempted to predict the interactions among multiple moving object using physical motion semantics, however, they did not account for realistic scenarios such as urban environments, which makes these networks better fitting for object linear motion in free space.

\textbf{Recurrent Neural Networks.}
Recently, Recurrent Neural Networks (RNN) have shown notable success in modeling data sequences and time-varying patterns. They organize in a recursively unfolded structure, which makes them a perfect choice for temporal analysis and sequence learning tasks, such as machine translation and human motion forecasting \cite{cho2014learning,bahdanau2014neural,song2017end,liu2016spatio,xue2018ss}.
Tree-structured RNN \cite{liu2016spatio}, illustrated spatio-temporal network organization analogous to \cite{jain2016structural}. However, their spatio-temporal architecture was designed around a skeletal-based human activity prediction such that, all the units had fixed dependencies and belong to one cohesive movement. This prior assumption does not fit with highly dynamic contexts such as crowd motion.

Few models \cite{jain2016structural,liang2016semantic,yuan2017temporal} structured RNN units based on graph topology that explicitly represented elements and their interactions semantics.
In our paper, we extend the generic spatio-temporal graph used in \cite{jain2016structural} in a hybrid manner, by combining globally-based human-human interaction with locality-based human-space interaction, in addition to using attention mechanism to distinctively model social interactions.



\section{Approach}
\subsection{Problem Definition}
Given a set of static objects $O$, and a set of pedestrians $ {V}$ and their trajectories $ {X_{v_i}^t}$ observed at time-steps t = 1,...,$T_{obs}$, our model predicts the future locations $ {\hat{X}_{v_i}^{t}}$ at t = $T_{obs}+1,....,T_{pred}$ time-steps, with regards to potential influence of any obstacles presented in the scene, such that $T_{obs}$ = 8,  $ {v_i} \in  {V}$, $T_{pred}$ = 12.
\subsection{Model Architecture}
\label{section:approach}

\begin{figure}[t]
\begin{center}
\subfloat[Crowded environment displayed over 2 time-steps.]{
\includegraphics[width=7cm, height=6cm]{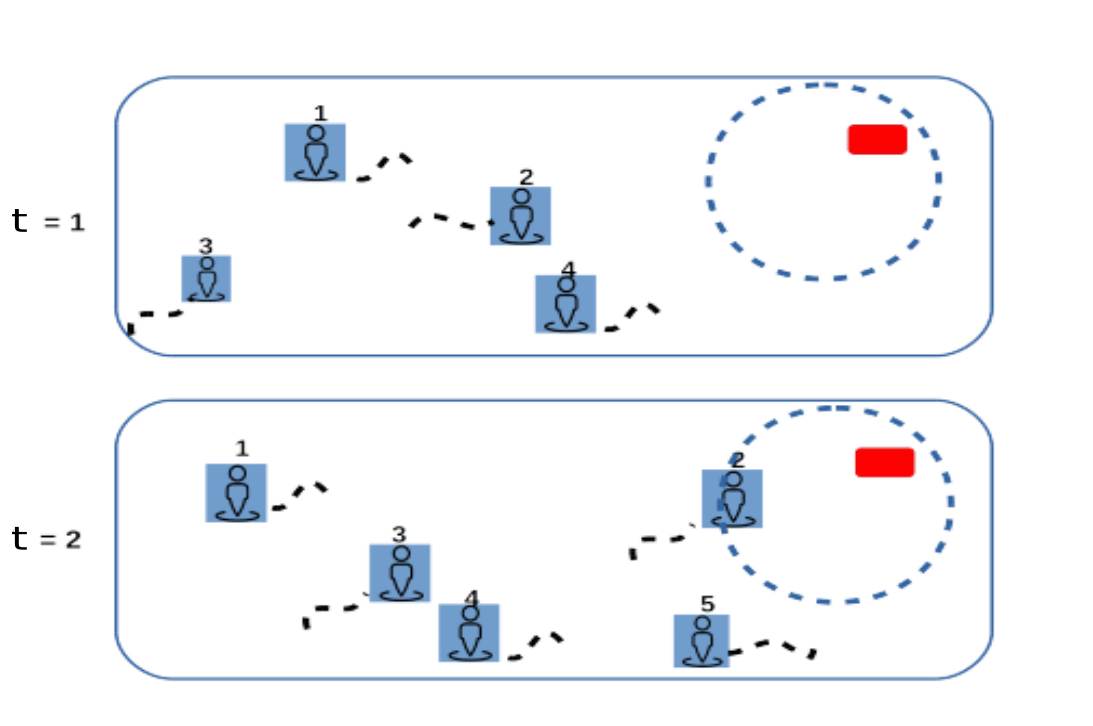}
}\hspace{3em}
\subfloat[Crowd mapping to abstract spatio-temporal graph unrolled through two time-steps]{\includegraphics[width=7cm, height=6cm]{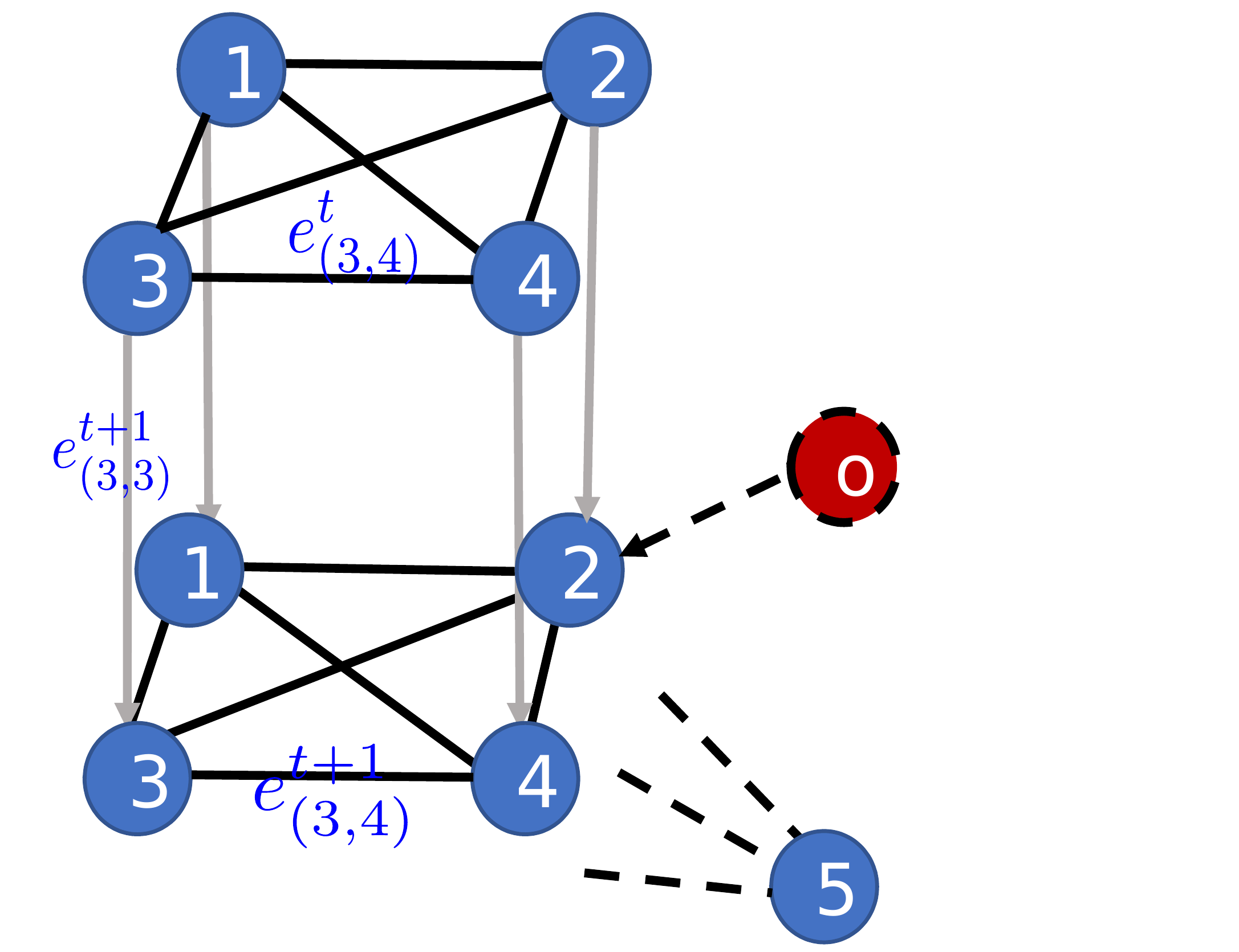}}
\end{center}

\caption{Crowd mapping to Spatio-temporal Graph. (a) A static obstacle is drawn as red rectangle surrounded by a virtual circle which indicates its neighborhood boundaries. (b) The Blue nodes represent pedestrians {1,2,3,4,5} and the red dashed node represents obstacle o such that $o \in O$. Directed downward lines indicate temporal edges linking the same node over time-steps and undirected lines are two-way spatial edges connecting pedestrian nodes. A directed edge is pointing from Obstacle node to pedestrian node to indicate obstacle influence on pedestrian. For the sake of clarity, we use dashed links from node (5) to indicate the remaining spatial edges. (Best viewed in color).} 
\label{fig:fig_1}
\end{figure}

The spatio-temporal graph is a dynamic structure that evolves temporally and spatially, due to the motion state of the pedestrians and changes in the scene (e.g. as the elements in the scene increase/decrease).
Figure \ref{fig:fig_1} shows the corresponding representation of crowd subjects in spatio-temporal graphs $\it{ {G}} = ( {V}, {\Sigma_S} , {\Sigma_T})$, comprising three key components: nodes set $ {V*}$, spatial edges set ${\Sigma_S}$ and temporal edges set ${\Sigma_T}$, where nodes represent the dynamic and the static element (e.g. pedestrians and static objects), spatial edges represent the relationship between two nodes to indicate the interaction between them. Temporal edges link the same pedestrian node over successive time-steps and thus connect the graph when it is unrolled over time.

Figure \ref{fig:fig_1}a illustrates the dynamic structure with an arbitrary crowd at two consecutive time-steps. At (t=1), there are four pedestrians. At (t=2), a new pedestrian (5) enters the scene. Notice that by (t=2), pedestrian (2) enters the vicinity of the red obstacle, where they appear to pass through the dashed circular boundary.
Figure \ref{fig:fig_1}b shows the corresponding spatio-temporal graph representation, which evolves dynamically over the spatial and temporal domain. This is evident when the graph unfolds at (t=2), where a new node is introduced for pedestrian (5) and all pedestrian nodes are connected by undirected edges to model the mutual interaction. This creates 2(N-1) spatial edges between pedestrian nodes at every time-step, where N is the number of pedestrians. In contrast, only a single directed edge is pointing from the obstacle node to the corresponding pedestrian node to depict the influence posed by the static obstacle on pedestrian (2).

The components of graph $\it{ {G}}$ are replaced with the corresponding LSTM components, {\it {temporal edgeLSTM}, \it {spatial edgeLSTM}, \it{nodeLSTM}}. 
The relationship between two nodes is characterized by their relative coordinates, where $x_ {v_2v_3}$ is the spatial distance between nodes $v_2$ and $v_3$, and $x_ {v_2v_2}$ is location of node v that changes over time.

Eq. (\ref{equ:embedded_spatial}) defines {\it {spatial edgeLSTM}} embedding function $\phi$ that takes as input: $x_ {v_2\textbf{.}}^t$, all the relative spatial distances between node $x_ {v_2}$ and its neighbors (e.g. including $x_ {v_2v_3}$), embedding weight matrix $W_s$ . 
\begin{gather}\label{equ:embedded_spatial}
e_ {v_2\textbf{.}}^t = \phi(x_ {v_2\textbf{.}}^t;W_s)\hspace{0.3em}
\end{gather}

The {\it {spatial edgeLSTM}}s take the embedded input feature along with previous spatial hidden states from all related nodes $h_ {v_2\textbf{.}}^{t-1}$ and transform them using normally initialized weight matrix $W^{lstm}_s$. The output hidden states vector $h_ {v_2\textbf{.}}^t$ is shown in 
Eq. (\ref{equ:hidden_spatial}).
\begin{gather}
 \label{equ:hidden_spatial}
h_ {v_2\textbf{.}}^t = LSTM(h_ {v_2\textbf{.}}^{t-1} , e_ {v_2\textbf{.}}^t , W^{lstm}_s)\hspace{0.3em}
\end{gather}

Eq. (\ref{equ:embedded_temporal}) defines {\it {temporal edgeLSTM}} embedding function $\phi$ that takes as input: the temporal location of pedestrian node $x_ {v_2v_2}^t$, embedding weight matrix $W_t$. 
\begin{gather}\label{equ:embedded_temporal}
e_ {v_2v_2}^t = \phi(x_ {v_2v_2}^t;W_t)\hspace{0.3em}
\end{gather}

Eq. (\ref{equ:hidden_temporal}) defines the LSTM cell and its inputs: previous temporal hidden state $h_ {v_2v_2}^{t-1}$, embedded input feature $e_ {v_2v_2}^t$ from Eq. (\ref{equ:embedded_temporal}) and  normally initialized weight matrix $W^{lstm}_t$ for transforming these inputs into the current hidden state $h_ {v_2v_2}^t$.

\begin{gather}
 \label{equ:hidden_temporal}
h_ {v_2v_2}^t = LSTM(h_ {v_2v_2}^{t-1} , e_ {v_2v_2}^t , W^{lstm}_t)\hspace{0.3em}
\end{gather}
\begin{figure*}[htp]
\centering
\includegraphics[width= 0.95\textwidth, height=6.5cm]{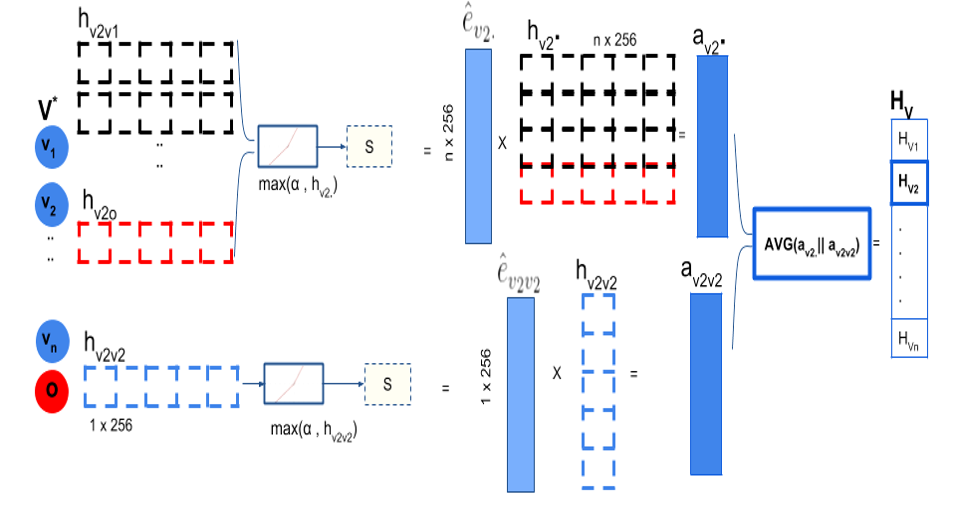}
\caption{ Multi-node attention mechanism pipeline for pedestrian node $v_2$ at time-step $t = 2$. On the left-most side, it shows nodes set $V^*$ = \{$V$,$O$\}. The black dashed vectors store $h_{v2\textbf{.}}^{t+1}$, the hidden states of the spatial edges related to node $v_2$. The red dashed vector stores $h_{v2o}^{t+1}$, the hidden state of spatial edge between node $v_2$ and obstacle $o$. The blue dashed vector stores $h_{v2v2}^{t+1}$, temporal edge hidden state of node $v2$. These hidden states are then passed into PReLU and Softmax (S) activations to generate new embeddings $\hat{e}$. The concrete blue vectors store spatial hidden states  $\hat{e}_{v_2\textbf{.}}^{t+1}$ and temporal hidden state $\hat{e}_{v_2v_2}^{t+1}$.  Multiplying the new embeddings vector by their hidden states array results in attention coefficients vectors $a$, where $a_{v_2\textbf{.}}^{t+1}$ is the spatial attention coefficients vector and $a_{v_2v_2}^{t+1}$ is the temporal attention coefficients vector.}
\label{fig:fig_2}
\end{figure*}


\subsection{Spatio-Temporal Attention Module}
Given the success of attentional mechanisms in sequence-based prediction of natural language processing applications, this work adopts the concept of attention-based generative algorithms  \cite{vaswani2017attention}. We propose a variation of Multi-Head method, a soft attention based on two simple operations, i.e. concatenation and averaging across all edge feature vectors for each node.
In \cite{vaswani2017attention}, the input comprises fixed number of words with fixed positions, and the Multi-Head attention works by stacking multiple attention layers (heads) in which each layer makes mappings between words in two sentences. 
We use a simple attention mechanism, i.e. Multi-Node attention, which only has a single layer that jointly pays attention to the features from spatial and temporal domains and store the attention coefficients into single vector for node $ {v_2}$ trajectory at each time-step. 
To illustrate this, Figure \ref{fig:fig_2} exemplifies attention on pedestrian (2) and its neighbors at time (t=2). Neighboring {\it{edgeLSTM}}s states are transformed before concatenation using the embedding function in Eq. (\ref{equ:prelu_all_edges}) and Eq. (\ref{equ:prelu_temp_edges}), which is a composite of Parametric ReLU and softmax. This combined activation ensures that hidden states remain within a small range of [-1,1] which will be mapped once again at the sampling stage to a range of normalized outputs range of [0,1].

\begin{equation}\label{equ:prelu_all_edges}
\hat{e}_{v_2\textbf{.}}^t = softmax(PReLU(h_{v_2\textbf{.}}^t))\hspace{0.3em}
\end{equation}
\begin{equation}\label{equ:prelu_temp_edges}
\hat{e}_{v_2v_2}^t =  softmax(PReLU(h_{v_2v_2}^t))\hspace{0.3em}
\end{equation} 

The Parametric ReLU as illustrated in Eq (\ref{equ:prelu_formula}), is the generalized ReLU function as it ties the leak parameter $\alpha$ as a network learnable parameter.
Employing such activation function with an adaptive leak parameters, allows a slightly different span of the negative hidden states along training batches. This has proved its benefit for the model prediction performance.
\begin{gather} \label{equ:prelu_formula}
PReLU(h) = max(0,h) + \alpha * min(0,h); \\ \nonumber  \alpha = 0.2
\end{gather}

The product of embedding vectors $\hat{e}$ with the original hidden states results in attention weights (also called coefficients). Eq. (\ref{equ:spatial_attention_coef}) and Eq. (\ref{equ:temporal_attention_coef}) shows the spatial attention coefficients $a_{v_2\textbf{.}}^t$ and temporal attention coefficients $a_{v_2v_2}^t$, respectively. 

\begin{equation}\label{equ:spatial_attention_coef}
 {a_{v_2\textbf{.}}^t} = \hat{e}_ {v_{2}\textbf{.}}^t h_ {v_{2}\textbf{.}}^t
\end{equation}
\begin{equation}\label{equ:temporal_attention_coef}
 {a_{v_2v_2}^t} = \hat{e}_ {v_2v_2}^t h_ {v_2v_2}^t 
\end{equation}

Eventually, these coefficients will be concatenated and averaged to generate the final weighted hidden states vector ${\it H_ {v_2}}^t$ as shown by Eq. (\ref{equ:attention_vec}):

\begin{equation}\label{equ:attention_vec}
{{\it{H}}_{v_2}^t} = \frac{\sum_{ {v}}^{N} ( {a_{v_2 v_2}^t} ||  {a_{v_2\textbf{.}}^t})}{N} ;\hspace{0.3em}
N = |\hspace{0.3em}  {a_{v_2 v_2}^t} || \hspace{0.3em}  {a_{v_2\textbf{.}}^t} \hspace{0.3em}|\hspace{0.3em}
\end{equation}

Comparing the Multi-Head attention with the single head multiplicative attention (scaled dot-Product), it turns out that the scaled dot-Product gives a compact representation of all incoming hidden states and it serves a similar objective to the linear pooling mechanism in \cite{alahi2016social} due to the highly variable-sized environment. However, it diminishes the expressive power lost upon compressing feature vectors size. 

While Multi-Head attention averages across the spatial and temporal attention coefficients without compressing their depth. Hence, we realized that retaining the vectors depth provides sufficient feature representation for learning the influence of pedestrians on each other.

The pedestrian location coordinates $x_ {v_2}^t$ are passed through an embedding layer $\phi$ as in Eq. (\ref{equ:embedded_node}) before its taken as input into \textit{nodeLSTM}: 
\begin{equation}\label{equ:embedded_node}
e_ {v_2}^t = \phi(x_ {v_2}^t;W_{embed})\hspace{0.3em}
\end{equation}

Finally, the output vector $ {{\it{H}}_{v_2}}^t$ is concatenated with previous hidden state $h_ {v_2}^{t-1}$, and  which are then passed to {\it{nodeLSTM}} $ {v_2}$,  along with transformation weight matrix $W^{lstm}$ to generate current hidden state $h_ {v_2}^t$.\newline

\begin{equation}\label{equ:nodelstm}
h_ {v_2}^t = LSTM(e_ {v_2}^t , concat(h_ {v_2}^{t-1} ,  {{\it{H}}_{v_2}}^t,e_ {v_2}^t ), W^{lstm})\hspace{0.3em}
\end{equation}

The future location of pedestrian is sampled from a bivariate normal distribution $\textit{N}$ as in Eq. (\ref{equ:sample}). For estimating the Mean $\mu$, variance $\sigma$ and correlation $\rho$ we apply a linear transformation layer in Eq. (\ref{equ:param_output}) $W_{out}$ to transform $h_ {v_2}^t$ into the estimated parameters.

\begin{equation}\label{equ:param_output}
(\mu_{ {v_2}}^{t+1},\sigma_{ {v_2}}^{t+1},\rho_{ {v_2}}^{t+1}) = W_{out}h_ {v_2}^t\hspace{0.3em}
\end{equation}
\begin{equation}\label{equ:sample}
(x_ {v_2}^{t+1},y_ {v_2}^{t+1}) \sim \textit{N}\hspace{0.3em}(\mu_{ {v_2}}^{t+1},\sigma_{ {v_2}}^{t+1},\rho_{ {v_2}}^{t+1})\hspace{0.3em}
\end{equation}

\section{Experimental Results}
\label{section:experiment}
\subsection{Datasets and Metrics}
Our evaluation is based on two widely-used datasets, ETH Walking Pedestrians (EWAP) \cite{pellegrini2009you}, UCY Students and Zara \cite{lerner2007crowds}. In total, the datasets consist of five videos taken from outdoor surveillance cameras.
The datasets contain 2206 human trajectories, exhibiting different traits that range between straight linear and curvilinear motion splines. From our observations, ETH scenes consist of more straight trajectories with few social interactions as the video captures people motion at the university entrance, while UCY scenes display more scenarios pertaining to human-space interactions. For example, the UCY-ZARA datasets include pedestrians bending at the shop entrance,
while UCY-University scenes have more social interactions among standing groups. Furthermore, these cases in particular, increase the unpredictability of an individual path unless social and spatial contexts are taken into account.
In our experiments, two benchmark metrics are used, i.e. Averaged Displacement Error ({\it ADE}) and Final Displacement Error ({\it FDE}) of the TrajNet challenge \cite{sadeghiankosaraju2018trajnet}, for measuring Euclidean deviations ({\it {in meters}}) between predicted trajectory and actual trajectory.

{\it{ Averaged Displacement Error}}:
The mean average {\it {l2 distances}} between predicted trajectory coordinates $(\hat{x},\hat{y})$ and true trajectory $ {(x,y)}$ for all time-steps $i=(1,..,n)$ over $N$ pedestrian trajectories in the scene.
\begin{equation}
\label{equ:ade}
{\it ADE} = \frac{\sum_{j=1}^{N} \frac{\sum_{i=1}^{n}  \sqrt{ (\hat{x}_i^j - {x}_i^j)^2 + ({\hat{y}_i}^j - {y}_i^j)^2}} {n}}{N}\hspace{0.3em}
\end{equation}


{\it {Final Displacement Error}}:
The average {\it {l2 distance}} between the final predict step $(\hat{x}_n,\hat{y}_n)$ and the true step $ {(x_n,y_n)}$ over $j$ pedestrians trajectory, where $j=(1,...,N)$. 
\begin{equation} 
\label{equ:fde}
{\it FDE} = \frac{\sum_{j=1}^{N} \sqrt{ ({\hat{x}^j}_n - {x}^j_n)^2 + ({\hat{y}^j}_n - {y}^j_n)^2 }} {N}\hspace{0.3em}
\end{equation}


\subsection{Ablation Study}
We have performed an ablation study by dropping the scaled-dot attention module from Social Attention and restoring back original settings of Structural-RNN, to study the usefulness of dot-Product attention model. The comparison between the quantitative results of both baselines with our method, shows that the scaled dot-Product performance is lower than the Multi-Node mechanism performance for the 5 datasets in Table \ref{table:table_1}. On the other hand, the optimal choice of the human-obstacle connectivity threshold ${\lambda} = 0.5$ parameter, was determined empirically, based on the objective of lowering the Euclidean errors for both evaluation metrics.

\subsection{Training Setup}
We accumulated trajectory data for every pedestrian with {\it skip\_rate} = 10 frames to avoid overfitting the minimal changes in pedestrian trajectory. Each LSTM cell is of 256 depth. We transform data into normalized interpolated pixel coordinates within range [0,1]. In batch processing, we fixate the batch size {{\it batch\_size} = 24}, observation length $T_{obs}$ = 8 time-steps (3.2 seconds), prediction length $T_{pred}$ = 12 time-steps (4.8 seconds) and epochs {{\it epoch\_num} = 100}. After several hyper-parameter tunings, learning rate is set as {{\it lr} = 0.001} and optimizer algorithm is {Adam}. Activation function in attention layer is {Parametric ReLU}, initialized to negative slope {\it ${\alpha}$} = 0.20 and fractionally degraded throughout the training process.
The training objective is to minimize the negative log-likelihood loss of the ${\it i^{th}}$ trajectory from time-step $T_{obs + 1}$ to $T_{pred}$:

\begin{equation} 
L_i = \quad {-\sum^{T_{pred}}_{t=T_{obs} + 1} log ( {P}(x_i^t,y_i^t|\sigma_i^t,\mu_i^t, \rho_i^t))\hspace{0.3em} } \end{equation}

\subsection{Quantitative Results}
As illustrated in Table \ref{table:table_1}, we set up experiments to evaluate our proposed models, H-H and H-H-O, which stand for Human-Human and Human-Human-Obstacle respectively.
The table has two segments, the first 4 rows evaluate our model with graph-based baselines: Social Attention and Structural-RNN, while the next 3 rows evaluate our model with state-of-the-art models: Social-LSTM and Social GAN (SGAN).
Our attention mechanism for graphs improved prediction for human-human interaction and human-obstacle interaction over the other graph-based baselines: Social Attention and Structural-RNN. This is observed from the average errors under column (AVG) in Table \ref{table:table_1}. Comparing H-H-O with Social Attention, H-H-O achieves 55\% in the average of ADE and 61\% in the average of FDE in all datasets.
As Social GAN and Social-LSTM display the best trajectories produced by their models, we extracted the average of minimum errors pertaining only to the best predicted trajectories in H-H-O model. It can be observed that the minimum FDE is considerably lower than minimum FDE generated by SGAN model and Social LSTM, due to our model awareness of surrounding context. This has made predictions to be plausible and compliant with the environmental constraint. The Social GAN work shows several versions of their model, so we selected their best model version which is SGAN-20V for our comparison.
The most significant improvement is realized when comparing our model with Social GAN model, under the Hotel set with 93\% reduction in FDE. Furthermore, the Hotel scene contains more static elements such as trees and lamp posts as indicated in Figure \ref{fig:fig_1}. The second best improvement is realized when comparing our model with Social-LSTM model under the ETH set with 89\% reduction in FDE. The ETH dataset consists of a set of tightly coupled trajectories due to the crowd at the university entrance. This is a busy contextual point where pedestrians are mostly concerned about avoiding collisions with each others at the entrance site.
Additionally, our model performance yields 69\% reduction on FDE metric in Ucy-University, which proves that embedding information about physical structure of the scene and busy interaction points, refines the model understanding of pedestrian navigation in crowded sites and reduced the prediction errors in FDE, as our model was more capable of predicting the final step on a pedestrian trajectory.
From the previous table, it is noticeable that the ADE and FDE exhibit small discrepancies due to the accumulative nature in prediction errors. If the predicted path was entirely approximate to the ground-truth, the final predicted point will not have large error, but if the prediction was increasingly deviating along the ground-truth, this can impact the final point errors. This supports our quantitative results as being consistent and realistic. 


\setlength{\tabcolsep}{4pt}

\begin{table*}[t]
\setlength\tabcolsep{6pt}
\begin{center}
\caption{
Prediction errors ADE/FDE (in meters). Our results are averaged over 30 sampled  sequences of 12-steps length for every set under our method. For baselines errors, Social Attention results are obtained upon re-training their model, while Structural-RNN results are obtained upon manual implementation of their architecture in PyTorch. 
}
\label{table:table_1}
\begin{tabular}{l  c  c  c  c  c  c }
\bottomrule

\hline\hline\noalign{\smallskip}

Method&  ETH &  HOTEL &  ZARA1 &  ZARA2 &   UNIV &  AVG\\ 
\noalign{\smallskip}
\hline
\noalign{\smallskip}
      
      { Structural-RNN} & 2.72/4.60 & 0.85/1.35 & 1.05/2.20 & 1.60/3.50 & 1.45/3.00 & 1.53/2.93\\ 
    { Social Attention}&3.60/4.70 &0.79/1.44 & 1.30/2.66 & 0.95/2.05 & 1.00/2.14 & 1.53/3.52 \\
	
    { H-H}&\textbf{1.19}/\textbf{2.00} & \textbf{0.39}/0.96 & 0.55/1.56 & 0.58/1.50 & 0.74/1.89 & \textbf{0.69}/1.58 \\ 

{ H-H-O} & 1.24/2.35 & 0.48/\textbf{0.80} & \textbf{0.51}/\textbf{1.15} & \textbf{0.56}/\textbf{1.13} & \textbf{0.69}/\textbf{1.45} & 0.70/\textbf{1.38}\\

\hline\hline\noalign{\smallskip}
{ Social LSTM} & 1.09/2.35 & 0.79/1.76 & 0.47/1.00 & 0.56/1.17 & 0.67/1.40 & 0.72/1.54 \\ 
{ SGAN-20V} &\textbf{0.81}/1.52 & 0.72/1.61 & \textbf{0.34}/0.69 & \textbf{0.42}/0.84 & 0.60/1.26 & \textbf{0.58}/1.18\\ 
{ Minimum H-H-O} &0.96/\textbf{0.16} & \textbf{0.35}/\textbf{0.11} & 0.57/\textbf{0.30} & 0.58/\textbf{0.33} & \textbf{0.53}/\textbf{0.38} & 0.60/\textbf{0.26}\\


\hline\hline\noalign{\smallskip}



\end{tabular}
\end{center}
\end{table*}


\subsection{Qualitative Results}

\begin{figure*}[t]
\centering
\begin{tabular}{c}


\subfloat[H-H model - Hotel scene]{\label{hh_hotel}
\includegraphics[width=6.5cm , height=4.5cm]{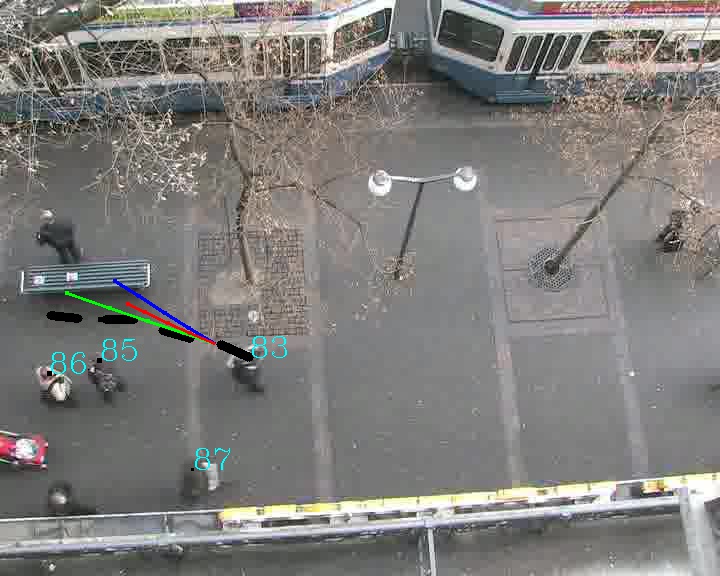}
}
\subfloat[H-H-O model - Hotel scene]{\label{hho_hotel}
\includegraphics[width=6.5cm , height=4.5cm]{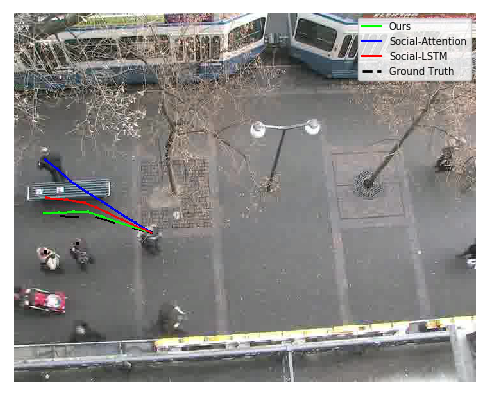}
}\\
\subfloat[H-H-O model - ZARA scene]{\label{hho_zara}
\includegraphics[width=6.5cm , height=4.5cm]{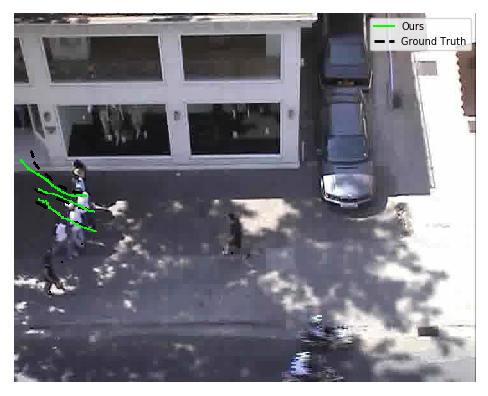}
}
\subfloat[H-H-O model - ZARA scene]{\label{hho_zara1}
\includegraphics[width=6.5cm , height=4.5cm]{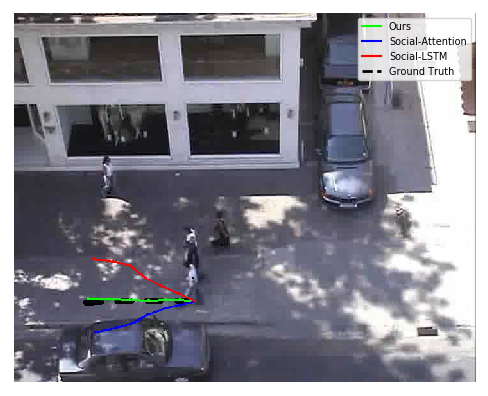}}
\end{tabular}

\caption{Visualization results for Hotel and Zara sets.}
\label{fig:fig_3}
\end{figure*}

In this section, we qualitatively evaluate model predictions in Hotel and ZARA sets. 
Figure \ref{fig:fig_3}
displays predicted paths from our models. We have spotted interesting cases for pedestrian moving near static objects, and compared both of our models outputs, Social Attention and Social LSTM with ground-truth trajectory. Notice the Human-Human model prediction for pedestrian walking near the bench in Figure \ref{hh_hotel}. The ground-truth shows that pedestrian is avoiding the bench, while Human-Human model spline achieves lower displacement than the baseline splines, those fail at evading the bench area. This case is correctly predicted in our Human-Human-Obstacle model as illustrated in Figure \ref{hho_hotel}. Additionally, Figure \ref{hho_zara1} shows that Social Attention and Social LSTM predicts plausible paths that pedestrian might have chosen, however, it is not compliant with pedestrian surrounding objects.
Thus, with the aid of obstacle awareness, our model understands pattern of collision avoidance with any static subject in their way. 

Figure \ref{hho_zara} plots trajectories from H-H-O model where pedestrians are bending toward the shop entrance, and our model generates splines that approximate the curvy ground-truth trajectory, as the model learns the motion pattern at the entrance point. 




In some situations, the predictions do not perfectly match the ground-truth path, although the deviations are quite small. This situation also applies for the baseline models. Upon extensive visual comparisons for all frames in all datasets, we confirmed that the erroneous results and deviations of the proposed method are much fewer than those found in the baselines plots. Quantitatively, Euclidean deviations at the path endings have been reduced by up to 61\%, which identifies the improvements that we highlighted earlier.


\section{Conclusion}
\label{section:conclusion}
In this paper we have presented a new spatio-temporal graph that operates on the local and global contexts around pedestrian, for predicting their trajectory in outdoor environments. For an accurate modeling of human-human interactions and human-space interactions, we employ a simplified version of Multi-Head attention mechanism for accumulating the influence from spatial and temporal subspaces. Our attention mechanism consistently demonstrated improved prediction results over baseline methods, for groups as well as individual non-linear trajectories. 
\bibliographystyle{IEEEtran}
\bibliography{IEEEexample}

\bibliographystyle{./IEEEtran}
\bibliography{./IEEEabrv,./IEEEexample}


@article{johansson2008specification,
  title={Specification of a microscopic pedestrian model by evolutionary adjustment to video tracking data [C]},
  author={Johansson, Anders and Helbing, Dirk and Shukla, Pradyumn K},
  journal={Advances in Complex System{\copyright} World Scientific Publishing Company},
  volume={25},
  year={2008}
}

@inproceedings{xue2018ss,
  title={SS-LSTM: A Hierarchical LSTM Model for Pedestrian Trajectory Prediction},
  author={Xue, Hao and Huynh, Du Q and Reynolds, Mark},
  booktitle={2018 IEEE Winter Conference on Applications of Computer Vision (WACV)},
  pages={1186--1194},
  year={2018},
  organization={IEEE}
}


@inproceedings{xue2017bi,
  title={Bi-Prediction: Pedestrian Trajectory Prediction Based on Bidirectional LSTM Classification},
  author={Xue, Hao and Huynh, Du Q and Reynolds, Mark},
  booktitle={Digital Image Computing: Techniques and Applications (DICTA), 2017 International Conference on},
  pages={1--8},
  year={2017},
  organization={IEEE}
}


@inproceedings{gupta2018social,
  title={Social GAN: Socially Acceptable Trajectories with Generative Adversarial Networks},
  author={Gupta, Agrim and Johnson, Justin and Fei-Fei, Li and Savarese, Silvio and Alahi, Alexandre},
  booktitle={IEEE Conference on Computer Vision and Pattern Recognition (CVPR)},
  number={CONF},
  year={2018}
}


@article{koppula2016anticipating,
  title={Anticipating human activities using object affordances for reactive robotic response},
  author={Koppula, Hema S and Saxena, Ashutosh},
  journal={IEEE transactions on pattern analysis and machine intelligence},
  volume={38},
  number={1},
  pages={14--29},
  year={2016},
  publisher={IEEE}
}

@inproceedings{fouhey2014predicting,
  title={Predicting object dynamics in scenes},
  author={Fouhey, David F and Zitnick, C Lawrence},
  booktitle={Proceedings of the IEEE Conference on Computer Vision and Pattern Recognition},
  pages={2019--2026},
  year={2014}
}

@inproceedings{gong2011multi,
  title={Multi-hypothesis motion planning for visual object tracking},
  author={Gong, Haifeng and Sim, Jack and Likhachev, Maxim and Shi, Jianbo},
  booktitle={Computer Vision (ICCV), 2011 IEEE International Conference on},
  pages={619--626},
  year={2011},
  organization={IEEE}
}

@article{velickovic2017graph,
  title={Graph Attention Networks},
  author={Velickovic, Petar and Cucurull, Guillem and Casanova, Arantxa and Romero, Adriana and Lio, Pietro and Bengio, Yoshua},
  journal={stat},
  volume={1050},
  pages={20},
  year={2017}
}

@inproceedings{santoro2017simple,
  title={A simple neural network module for relational reasoning},
  author={Santoro, Adam and Raposo, David and Barrett, David G and Malinowski, Mateusz and Pascanu, Razvan and Battaglia, Peter and Lillicrap, Tim},
  booktitle={Advances in neural information processing systems},
  pages={4974--4983},
  year={2017}
}

@inproceedings{alahi2016social,
  title={Social lstm: Human trajectory prediction in crowded spaces},
  author={Alahi, Alexandre and Goel, Kratarth and Ramanathan, Vignesh and Robicquet, Alexandre and Fei-Fei, Li and Savarese, Silvio},
  booktitle={Proceedings of the IEEE Conference on Computer Vision and Pattern Recognition},
  pages={961--971},
  year={2016}
}

@inproceedings{jain2016structural,
  title={Structural-RNN: Deep learning on spatio-temporal graphs},
  author={Jain, Ashesh and Zamir, Amir R and Savarese, Silvio and Saxena, Ashutosh},
  booktitle={Proceedings of the IEEE Conference on Computer Vision and Pattern Recognition},
  pages={5308--5317},
  year={2016}
}

@article{graves2013generating,
  title={Generating sequences with recurrent neural networks},
  author={Graves, Alex},
  journal={arXiv preprint arXiv:1308.0850},
  year={2013}
}


@article{goel2015learning,
  title={Learning Causalities behind Human Trajectories},
  author={Goel, Kratarth and Robicquet, Alexandre},
  year={2015}
}


@article{helbing1995social,
  title={Social force model for pedestrian dynamics},
  author={Helbing, Dirk and Molnar, Peter},
  journal={Physical review E},
  volume={51},
  number={5},
  pages={4282},
  year={1995},
  publisher={APS}
}

@inproceedings{kitani2012activity,
  title={Activity forecasting},
  author={Kitani, Kris M and Ziebart, Brian D and Bagnell, James Andrew and Hebert, Martial},
  booktitle={European Conference on Computer Vision},
  pages={201--214},
  year={2012},
  organization={Springer}
}

@inproceedings{kretzschmar2014learning,
  title={Learning to predict trajectories of cooperatively navigating agents},
  author={Kretzschmar, Henrik and Kuderer, Markus and Burgard, Wolfram},
  booktitle={Robotics and Automation (ICRA), 2014 IEEE International Conference on},
  pages={4015--4020},
  year={2014},
  organization={IEEE}
}
@article{li2015gated,
  title={Gated graph sequence neural networks},
  author={Li, Yujia and Tarlow, Daniel and Brockschmidt, Marc and Zemel, Richard},
  journal={arXiv preprint arXiv:1511.05493},
  year={2015}
}

@article{makris2005learning,
  title={Learning semantic scene models from observing activity in visual surveillance},
  author={Makris, Dimitrios and Ellis, Tim},
  journal={IEEE Transactions on Systems, Man, and Cybernetics, Part B (Cybernetics)},
  volume={35},
  number={3},
  pages={397--408},
  year={2005},
  publisher={IEEE}
}

@article{morris2008survey,
  title={A survey of vision-based trajectory learning and analysis for surveillance},
  author={Morris, Brendan Tran and Trivedi, Mohan Manubhai},
  journal={IEEE transactions on circuits and systems for video technology},
  volume={18},
  number={8},
  pages={1114--1127},
  year={2008},
  publisher={IEEE}
}

@inproceedings{pellegrini2009you,
  title={You'll never walk alone: Modeling social behavior for multi-target tracking},
  author={Pellegrini, Stefano and Ess, Andreas and Schindler, Konrad and Van Gool, Luc},
  booktitle={Computer Vision, 2009 IEEE 12th International Conference on},
  pages={261--268},
  year={2009},
  organization={IEEE}
}

@inproceedings{robicquet2016learning,
  title={Learning social etiquette: Human trajectory understanding in crowded scenes},
  author={Robicquet, Alexandre and Sadeghian, Amir and Alahi, Alexandre and Savarese, Silvio},
  booktitle={European conference on computer vision},
  pages={549--565},
  year={2016},
  organization={Springer}
}

@article{scarselli2009graph,
  title={The graph neural network model},
  author={Scarselli, Franco and Gori, Marco and Tsoi, Ah Chung and Hagenbuchner, Markus and Monfardini, Gabriele},
  journal={IEEE Transactions on Neural Networks},
  volume={20},
  number={1},
  pages={61--80},
  year={2009},
  publisher={IEEE}
}

@conference{vemula2018social,
author = {Anirudh Vemula and Katharina Muelling and Jean Oh},
title = {Social Attention: Modeling Attention in Human Crowds},
booktitle = {Proceedings of the International Conference on Robotics and Automation (ICRA) 2018},
year = {2018},
month = {May},
} 

@article{varshneya2017human,
  title={Human trajectory prediction using spatially aware deep attention models},
  author={Varshneya, Daksh and Srinivasaraghavan, G},
  journal={arXiv preprint arXiv:1705.09436},
  year={2017}
}

@article{cho2014learning,
  title={Learning phrase representations using RNN encoder-decoder for statistical machine translation},
  author={Cho, Kyunghyun and Van Merri{\"e}nboer, Bart and Gulcehre, Caglar and Bahdanau, Dzmitry and Bougares, Fethi and Schwenk, Holger and Bengio, Yoshua},
  journal={arXiv preprint arXiv:1406.1078},
  year={2014}
}

@article{bahdanau2014neural,
  title={Neural machine translation by jointly learning to align and translate},
  author={Bahdanau, Dzmitry and Cho, Kyunghyun and Bengio, Yoshua},
  journal={arXiv preprint arXiv:1409.0473},
  year={2014}
}

@inproceedings{song2017end,
  title={An End-to-End Spatio-Temporal Attention Model for Human Action Recognition from Skeleton Data.},
  author={Song, Sijie and Lan, Cuiling and Xing, Junliang and Zeng, Wenjun and Liu, Jiaying},
  booktitle={AAAI},
  volume={1},
  number={2},
  pages={7},
  year={2017}
}

@inproceedings{sutskever2014sequence,
  title={Sequence to sequence learning with neural networks},
  author={Sutskever, Ilya and Vinyals, Oriol and Le, Quoc V},
  booktitle={Advances in neural information processing systems},
  pages={3104--3112},
  year={2014}
}

@article{hochreiter1997long,
  title={Long short-term memory},
  author={Hochreiter, Sepp and Schmidhuber, J{\"u}rgen},
  journal={Neural computation},
  volume={9},
  number={8},
  pages={1735--1780},
  year={1997},
  publisher={MIT Press}
}

@inproceedings{liu2016spatio,
  title={Spatio-temporal lstm with trust gates for 3d human action recognition},
  author={Liu, Jun and Shahroudy, Amir and Xu, Dong and Wang, Gang},
  booktitle={European Conference on Computer Vision},
  pages={816--833},
  year={2016},
  organization={Springer}
}
@inproceedings{lerner2007crowds,
  title={Crowds by example},
  author={Lerner, Alon and Chrysanthou, Yiorgos and Lischinski, Dani},
  booktitle={Computer Graphics Forum},
  volume={26},
  number={3},
  pages={655--664},
  year={2007},
  organization={Wiley Online Library}
}

@inproceedings{vaswani2017attention,
  title={Attention is all you need},
  author={Vaswani, Ashish and Shazeer, Noam and Parmar, Niki and Uszkoreit, Jakob and Jones, Llion and Gomez, Aidan N and Kaiser, {\L}ukasz and Polosukhin, Illia},
  booktitle={Advances in Neural Information Processing Systems},
  pages={6000--6010},
  year={2017}
}

@inproceedings{karasev2016intent,
  title={Intent-aware long-term prediction of pedestrian motion},
  author={Karasev, Vasiliy and Ayvaci, Alper and Heisele, Bernd and Soatto, Stefano},
  booktitle={Robotics and Automation (ICRA), 2016 IEEE International Conference on},
  pages={2543--2549},
  year={2016},
  organization={IEEE}
}

@article{lee2017desire,
  title={Desire: Distant future prediction in dynamic scenes with interacting agents},
  author={Lee, Namhoon and Choi, Wongun and Vernaza, Paul and Choy, Christopher B and Torr, Philip HS and Chandraker, Manmohan},
  year={2017}
}

@inproceedings{zhou2016learning,
  title={Learning time series models for pedestrian motion prediction},
  author={Zhou, Chenghui and Balle, Borja and Pineau, Joelle},
  booktitle={Robotics and Automation (ICRA), 2016 IEEE International Conference on},
  pages={3323--3330},
  year={2016},
  organization={IEEE}
}
@inproceedings{ellis2009modelling,
  title={Modelling pedestrian trajectory patterns with gaussian processes},
  author={Ellis, David and Sommerlade, Eric and Reid, Ian},
  booktitle={Computer Vision Workshops (ICCV Workshops), 2009 IEEE 12th International Conference on},
  pages={1229--1234},
  year={2009},
  organization={IEEE}
}

@inproceedings{kim2011gaussian,
  title={Gaussian process regression flow for analysis of motion trajectories},
  author={Kim, Kihwan and Lee, Dongryeol and Essa, Irfan},
  booktitle={Computer vision (ICCV), 2011 IEEE international conference on},
  pages={1164--1171},
  year={2011},
  organization={IEEE}
}

@inproceedings{ziebart2009planning,
  title={Planning-based prediction for pedestrians},
  author={Ziebart, Brian D and Ratliff, Nathan and Gallagher, Garratt and Mertz, Christoph and Peterson, Kevin and Bagnell, J Andrew and Hebert, Martial and Dey, Anind K and Srinivasa, Siddhartha},
  booktitle={Intelligent Robots and Systems, 2009. IROS 2009. IEEE/RSJ International Conference on},
  pages={3931--3936},
  year={2009},
  organization={IEEE}
}

@article{hu2004hierarchical,
  title={A hierarchical self-organizing approach for learning the patterns of motion trajectories},
  author={Hu, Weiming and Xie, Dan and Tan, Tieniu},
  journal={IEEE Transactions on Neural Networks},
  volume={15},
  number={1},
  pages={135--144},
  year={2004},
  publisher={IEEE}
}

@inproceedings{goldhammer2015camera,
  title={Camera based pedestrian path prediction by means of polynomial least-squares approximation and multilayer perceptron neural networks},
  author={Goldhammer, Michael and K{\"o}hler, Sebastian and Doll, Konrad and Sick, Bernhard},
  booktitle={SAI Intelligent Systems Conference (IntelliSys), 2015},
  pages={390--399},
  year={2015},
  organization={IEEE}
}

@article{yuan2017temporal,
  title={Temporal dynamic graph LSTM for action-driven video object detection},
  author={Yuan, Yuan and Liang, Xiaodan and Wang, Xiaolong and Yeung, Dit Yan and Gupta, Abhinav},
  journal={arXiv preprint arXiv:1708.00666},
  year={2017}
}


@inproceedings{liang2016semantic,
  title={Semantic object parsing with graph lstm},
  author={Liang, Xiaodan and Shen, Xiaohui and Feng, Jiashi and Lin, Liang and Yan, Shuicheng},
  booktitle={European Conference on Computer Vision},
  pages={125--143},
  year={2016},
  organization={Springer}
}

@inproceedings{mehran2009abnormal,
  title={Abnormal crowd behavior detection using social force model},
  author={Mehran, Ramin and Oyama, Alexis and Shah, Mubarak},
  booktitle={Computer Vision and Pattern Recognition, 2009. CVPR 2009. IEEE Conference on},
  pages={935--942},
  year={2009},
  organization={IEEE}
}

@article{zanlungo2011social,
  title={Social force model with explicit collision prediction},
  author={Zanlungo, Francesco and Ikeda, Tetsushi and Kanda, Takayuki},
  journal={EPL (Europhysics Letters)},
  volume={93},
  number={6},
  pages={68005},
  year={2011},
  publisher={IOP Publishing}
}

@inproceedings{yamaguchi2011you,
  title={Who are you with and where are you going?},
  author={Yamaguchi, Kota and Berg, Alexander C and Ortiz, Luis E and Berg, Tamara L},
  booktitle={Computer Vision and Pattern Recognition (CVPR), 2011 IEEE Conference on},
  pages={1345--1352},
  year={2011},
  organization={IEEE}
}

@article{graves2005framewise,
  title={Framewise phoneme classification with bidirectional LSTM and other neural network architectures},
  author={Graves, Alex and Schmidhuber, J{\"u}rgen},
  journal={Neural Networks},
  volume={18},
  number={5-6},
  pages={602--610},
  year={2005},
  publisher={Elsevier}
}

@article{bartoli2017context,
  title={Context-aware trajectory prediction},
  author={Bartoli, Federico and Lisanti, Giuseppe and Ballan, Lamberto and Del Bimbo, Alberto},
  journal={arXiv preprint arXiv:1705.02503},
  year={2017}
}

@inproceedings{battaglia2016interaction,
  title={Interaction networks for learning about objects, relations and physics},
  author={Battaglia, Peter and Pascanu, Razvan and Lai, Matthew and Rezende, Danilo Jimenez and others},
  booktitle={Advances in neural information processing systems},
  pages={4502--4510},
  year={2016}
}

@article{van2018relational,
  title={Relational neural expectation maximization: Unsupervised discovery of objects and their interactions},
  author={van Steenkiste, Sjoerd and Chang, Michael and Greff, Klaus and Schmidhuber, J{\"u}rgen},
  journal={arXiv preprint arXiv:1802.10353},
  year={2018}
}

@article{xie2018learning,
  title={Learning and Inferring “Dark Matter” and Predicting Human Intents and Trajectories in Videos},
  author={Xie, Dan and Shu, Tianmin and Todorovic, Sinisa and Zhu, Song-Chun},
  journal={IEEE transactions on pattern analysis and machine intelligence},
  volume={40},
  number={7},
  pages={1639--1652},
  year={2018},
  publisher={IEEE}
}

@article{Kipf2017Semi-supervised,
title = {Semi-supervised classification with graph convolutional networks},
author = {Kipf, T. N. and Welling, M.},
journal = {International Conference on Learning Representations (ICLR)},
year = {2017},
}

@article{sadeghiankosaraju2018trajnet,
  title={TrajNet: Towards a Benchmark for Human Trajectory Prediction},
  author={Sadeghian, Amir and Kosaraju, Vineet and Gupta, Agrim and Savarese, Silvio and Alahi, Alexandre},
  journal={arXiv preprint},
  volume={},
  number={},
  pages={},
  year={2018}
}


@article{fernando2018soft+,
  title={Soft+ hardwired attention: An lstm framework for human trajectory prediction and abnormal event detection},
  author={Fernando, Tharindu and Denman, Simon and Sridharan, Sridha and Fookes, Clinton},
  journal={Neural networks},
  volume={108},
  pages={466--478},
  year={2018},
  publisher={Elsevier}
}




An example of a IEEEtran control entry which can change some IEEEtran.bst
settings. An entry like this must be cited via \bstctlcite{} command
before the first real \cite{}. The same entry key cannot be called twice
(just like multiple \cite{} of the same entry key place only one entry
in the bibliography.)
The available control fields are:

CTLuse_article_number
"no" turns off the display of the number for articles.
"yes" enables

CTLuse_paper
"no" turns off the display of the paper and type fields in inproceedings.
"yes" enables

CTLuse_url
"no" turns off the display of the url field.
"yes" enables

CTLuse_forced_etal 
"no" turns off the forced use of "et al."
"yes" enables

CTLmax_names_forced_etal
The maximum number of names that can be present beyond which an "et al."
usage is forced. Be sure that CTLnames_show_etal (below)
is not greater than this value!

CTLnames_show_etal
The number of names that will be shown with a forced "et al.".
Must be less than or equal to CTLmax_names_forced_etal

CTLuse_alt_spacing 
"no" turns off the alternate interword spacing for entries with URLs.
"yes" enables
Will not have any effect (is disabled) if the display of urls is disabled.

CTLalt_stretch_factor
If alternate interword spacing for entries with URLs is enabled, this is
the interword spacing stretch factor that will be used. For example, the
default "4" here means that the interword spacing in entries with URLs can
stretch to four times normal. Does not have to be an integer.

CTLdash_repeated_names
"no" turns off the "dashification" of repeated (i.e., identical to those
of the previous entry) names. The IEEE normally does this.
"yes" enables

CTLname_format_string
The name format control string as explained in the BibTeX style hacking
guide.
IEEE style "{f.~}{vv~}{ll}{, jj}" is the default,

CTLname_latex_cmd
A LaTeX command that each name will be fed to (e.g., "\textsc").
Leave empty if no special font is desired for the names.
The default is empty.

CTLname_url_prefix
The prefix text used before URLs (if they are enabled).
The default is "[Online]. Available:" A space will be inserted after this
text. If this space is not wanted, just use \relax at the end of the
prefix text.


Those fields that are not to be changed can be left out.
@IEEEtranBSTCTL{IEEEexample:BSTcontrol,
  CTLuse_article_number     = "yes",
  CTLuse_paper              = "yes",
  CTLuse_url                = "yes",
  CTLuse_forced_etal        = "no",
  CTLmax_names_forced_etal  = "10",
  CTLnames_show_etal        = "1",
  CTLuse_alt_spacing        = "yes",
  CTLalt_stretch_factor     = "4",
  CTLdash_repeated_names    = "yes",
  CTLname_format_string     = "{f.~}{vv~}{ll}{, jj}",
  CTLname_latex_cmd         = "",
  CTLname_url_prefix        = "[Online]. Available:"
}





IEEEfull.bib
V1.14 (2015/08/28)
Copyright (c) 2002-2015 by Michael Shell
See: http://www.michaelshell.org/
for current contact information.

BibTeX bibliography string definitions of the FULL titles of
IEEE journals and magazines and online publications.

This file is designed for bibliography styles that require 
full-length titles and is not for use in bibliographies that
abbreviate titles.

Support sites:
http://www.michaelshell.org/tex/ieeetran/
http://www.ctan.org/pkg/ieeetran
and/or
http://www.ieee.org/

Special thanks to Laura Hyslop, Ken Rawson, Kevin Lisankie and
Mona Mittra of the IEEE for their help in obtaining the information needed
to compile this file. Also, Volker Kuhlmann, Moritz Borgmann,
Yannick Berker, Nicol�s Barabino, Chuanren Wu and Santiago Cogollos Borras
kindly provided some corrections and additions.


*************************************************************************
Legal Notice:
This code is offered as-is without any warranty either expressed or
implied; without even the implied warranty of MERCHANTABILITY or
FITNESS FOR A PARTICULAR PURPOSE! 
User assumes all risk.
In no event shall the IEEE or any contributor to this code be liable for
any damages or losses, including, but not limited to, incidental,
consequential, or any other damages, resulting from the use or misuse
of any information contained here.

All comments are the opinions of their respective authors and are not
necessarily endorsed by the IEEE.

This work is distributed under the LaTeX Project Public License (LPPL)
( http://www.latex-project.org/ ) version 1.3, and may be freely used,
distributed and modified. A copy of the LPPL, version 1.3, is included
in the base LaTeX documentation of all distributions of LaTeX released
2003/12/01 or later.
Retain all contribution notices and credits.
** Modified files should be clearly indicated as such, including  **
** renaming them and changing author support contact information. **
*************************************************************************


USAGE:

\bibliographystyle{mybstfile}
\bibliography{IEEEfull,mybibfile}

where the IEEE titles in the .bib database entries use the strings
defined here. e.g.,


   journal = IEEE_J_AC,


to yield "{IEEE} Transactions on Automatic Control"


WARNING: The IEEE uses abbreviated journal titles in their bibliographies!
Because this file provides the full titles, you should NOT use this file
for work that is to be submitted to the IEEE.

For IEEE work, you should use the abbreviated titles provided in the
companion file, IEEEabrv.bib.


** NOTES **

 1. Journals have been grouped according to subject in order to make it
    easier to locate and extract the definitions for related journals - 
    as most works use references that are confined to a single topic.
    Magazines are listed in straight alphabetical order.

 2. String names are closely based on IEEE's own internal acronyms.

 3. Older, out-of-print IEEE titles are included (but not including titles
    dating prior to the IEEE's formation from the IRE and AIEE in 1963).






IEEE Journals 


aerospace and military
@STRING{IEEE_J_ANNE       = "{IEEE} Transactions on Aeronautical and Navigational Electronics"}
@STRING{IEEE_J_AES        = "{IEEE} Transactions on Aerospace and Electronic Systems"}
@STRING{IEEE_J_ANE        = "{IEEE} Transactions on Aerospace and Navigational Electronics"}
@STRING{IEEE_J_AS         = "{IEEE} Transactions on Aerospace"}
@STRING{IEEE_J_AIRE       = "{IEEE} Transactions on Airborne Electronics"}
@STRING{IEEE_J_MIL        = "{IEEE} Transactions on Military Electronics"}



autos, transportation and vehicles (non-aerospace)
@STRING{IEEE_J_ITS        = "{IEEE} Transactions on Intelligent Transportation Systems"}
@STRING{IEEE_J_IV         = "{IEEE} Transactions on Intelligent Vehicles"}
@STRING{IEEE_J_TTE        = "{IEEE} Transactions on Transportation Electrification"}
@STRING{IEEE_J_VC         = "{IEEE} Transactions on Vehicular Communications"}
@STRING{IEEE_J_VT         = "{IEEE} Transactions on Vehicular Technology"}



circuits, signals, systems, audio and controls
@STRING{IEEE_J_STSP       = "{IEEE} Journal of Selected Topics in Signal Processing"}
@STRING{IEEE_J_SPL        = "{IEEE} Signal Processing Letters"}
@STRING{IEEE_J_SYST       = "{IEEE} Systems Journal"}
@STRING{IEEE_J_ASSP       = "{IEEE} Transactions on Acoustics, Speech, and Signal Processing"}
@STRING{IEEE_J_AU         = "{IEEE} Transactions on Audio"}
@STRING{IEEE_J_AUEA       = "{IEEE} Transactions on Audio and Electroacoustics"}
in 2014 ASL became ASLP
@STRING{IEEE_J_ASLP       = "{IEEE/ACM} Transactions on Audio, Speech, and Language Processing"}
@STRING{IEEE_J_ASL        = "{IEEE} Transactions on Audio, Speech, and Language Processing"}
@STRING{IEEE_J_AC         = "{IEEE} Transactions on Automatic Control"}
@STRING{IEEE_J_CAS        = "{IEEE} Transactions on Circuits and Systems"}
@STRING{IEEE_J_CASVT      = "{IEEE} Transactions on Circuits and Systems for Video Technology"}
@STRING{IEEE_J_CASI       = "{IEEE} Transactions on Circuits and Systems---Part {I}: Fundamental Theory and Applications"}
@STRING{IEEE_J_CASII      = "{IEEE} Transactions on Circuits and Systems---Part {II}: Analog and Digital Signal Processing"}
in 2004 CASI and CASII renamed part title to CASI_RP and CASII_EB, respectively
@STRING{IEEE_J_CASI_RP    = "{IEEE} Transactions on Circuits and Systems---Part {I}: Regular Papers"}
@STRING{IEEE_J_CASII_EB   = "{IEEE} Transactions on Circuits and Systems---Part {II}: Express Briefs"}
@STRING{IEEE_J_CT         = "{IEEE} Transactions on Circuit Theory"}
@STRING{IEEE_J_CST        = "{IEEE} Transactions on Control Systems Technology"}
@STRING{IEEE_J_ETCAS      = "{IEEE} Transactions on Emerging and Selected Topics in Circuits and Systems"}
@STRING{IEEE_J_SP         = "{IEEE} Transactions on Signal Processing"}
@STRING{IEEE_J_SU         = "{IEEE} Transactions on Sonics and Ultrasonics"}
@STRING{IEEE_J_SAP        = "{IEEE} Transactions on Speech and Audio Processing"}
@STRING{IEEE_J_UE         = "{IEEE} Transactions on Ultrasonics Engineering"}
@STRING{IEEE_J_UFFC       = "{IEEE} Transactions on Ultrasonics, Ferroelectrics, and Frequency Control"}



communications
@STRING{IEEE_J_COML       = "{IEEE} Communications Letters"}
@STRING{IEEE_J_JSAC       = "{IEEE} Journal on Selected Areas in Communications"}
@STRING{IEEE_J_COM        = "{IEEE} Transactions on Communications"}
@STRING{IEEE_J_COMT       = "{IEEE} Transactions on Communication Technology"}
@STRING{IEEE_J_WCOM       = "{IEEE} Transactions on Wireless Communications"}
@STRING{IEEE_J_WCOML      = "{IEEE} Wireless Communications Letters"}



components, packaging and manufacturing
@STRING{IEEE_J_ADVP       = "{IEEE} Transactions on Advanced Packaging"}
@STRING{IEEE_J_CHMT       = "{IEEE} Transactions on Components, Hybrids and Manufacturing Technology"}
in 2011 CAPT became CPMT
@STRING{IEEE_J_CPMT       = "{IEEE} Transactions on Components, Packaging and Manufacturing Technology"}
@STRING{IEEE_J_CPMTA      = "{IEEE} Transactions on Components, Packaging and Manufacturing Technology---Part {A}"}
@STRING{IEEE_J_CPMTB      = "{IEEE} Transactions on Components, Packaging and Manufacturing Technology---Part {B}: Advanced Packaging"}
@STRING{IEEE_J_CPMTC      = "{IEEE} Transactions on Components, Packaging and Manufacturing Technology---Part {C}: Manufacturing"}
@STRING{IEEE_J_CAPTS      = "{IEEE} Transactions on Components and Packaging Technologies"}
@STRING{IEEE_J_CAPT       = "{IEEE} Transactions on Components and Packaging Technology"}
@STRING{IEEE_J_CPART      = "{IEEE} Transactions on Component Parts"}
@STRING{IEEE_J_EPM        = "{IEEE} Transactions on Electronics Packaging Manufacturing"}
@STRING{IEEE_J_MFT        = "{IEEE} Transactions on Manufacturing Technology"}
@STRING{IEEE_J_PHP        = "{IEEE} Transactions on Parts, Hybrids and Packaging"}
@STRING{IEEE_J_PMP        = "{IEEE} Transactions on Parts, Materials and Packaging"}



CAD
@STRING{IEEE_J_TCAD       = "{IEEE} Journal on Technology in Computer Aided Design"}
@STRING{IEEE_J_CAD        = "{IEEE} Transactions on Computer-Aided Design of Integrated Circuits and Systems"}



coding, data, information, knowledge
@STRING{IEEE_J_BD         = "{IEEE} Transactions on Big Data"}
@STRING{IEEE_J_IFS        = "{IEEE} Transactions on Information Forensics and Security"}
@STRING{IEEE_J_IT         = "{IEEE} Transactions on Information Theory"}
@STRING{IEEE_J_KDE        = "{IEEE} Transactions on Knowledge and Data Engineering"}



computers, computation, networking and software
@STRING{IEEE_J_CAL        = "{IEEE} Computer Architecture Letters"}
@STRING{IEEE_J_ES         = "{IEEE} Embedded Systems Letters"}
@STRING{IEEE_J_IOT        = "{IEEE} Internet of Things Journal"}
@STRING{IEEE_J_XCDC       = "{IEEE} Journal on Exploratory Solid-State Computational Devices and Circuits"}
@STRING{IEEE_J_MMCT       = "{IEEE} Journal on Multiscale and Multiphysics Computational Techniques"}
@STRING{IEEE_J_SUSC       = "{IEEE} Sustainable Computing"}
@STRING{IEEE_J_CC         = "{IEEE} Transactions on Cloud Computing"}
@STRING{IEEE_J_CSS        = "{IEEE} Transactions on Computational Social Systems"}
@STRING{IEEE_J_C          = "{IEEE} Transactions on Computers"}
@STRING{IEEE_J_CNS        = "{IEEE} Transactions on Control of Network Systems"}
@STRING{IEEE_J_DSC        = "{IEEE} Transactions on Dependable and Secure Computing"}
@STRING{IEEE_J_ECOMP      = "{IEEE} Transactions on Electronic Computers"}
@STRING{IEEE_J_ETC        = "{IEEE} Transactions on Emerging Topics in Computing"}
@STRING{IEEE_J_EVC        = "{IEEE} Transactions on Evolutionary Computation"}
@STRING{IEEE_J_FUZZ       = "{IEEE} Transactions on Fuzzy Systems"}
@STRING{IEEE_J_MC         = "{IEEE} Transactions on Mobile Computing"}
@STRING{IEEE_J_MSCS       = "{IEEE} Transactions on Multi-Scale Computing Systems"}
@STRING{IEEE_J_NET        = "{IEEE/ACM} Transactions on Networking"}
@STRING{IEEE_J_NSE        = "{IEEE} Transactions on Network Science and Engineering"}
@STRING{IEEE_J_NSM        = "{IEEE} Transactions on Network and Service Management"}
@STRING{IEEE_J_NN         = "{IEEE} Transactions on Neural Networks"}
in 2012 NN became NNLS
@STRING{IEEE_J_NNLS       = "{IEEE} Transactions on Neural Networks and Learning Systems"}
@STRING{IEEE_J_PDS        = "{IEEE} Transactions on Parallel and Distributed Systems"}
@STRING{IEEE_J_SC         = "{IEEE} Transactions on Services Computing"}
@STRING{IEEE_J_SIPN       = "{IEEE} Transactions on Signal and Information Processing over Networks"}
@STRING{IEEE_J_SE         = "{IEEE} Transactions on Software Engineering"}



computer graphics, imaging, and multimedia
@STRING{IEEE_J_JDT        = "{IEEE/OSA} Journal of Display Technology"}
@STRING{IEEE_J_IP         = "{IEEE} Transactions on Image Processing"}
@STRING{IEEE_J_MM         = "{IEEE} Transactions on Multimedia"}
@STRING{IEEE_J_VCG        = "{IEEE} Transactions on Visualization and Computer Graphics"}



cybernetics, ergonomics, robots, man-machine, artificial intelligence and automation
@STRING{IEEE_J_JAS        = "{IEEE/CAA} Journal of Automatica Sinica"}
@STRING{IEEE_J_JRA        = "{IEEE} Journal of Robotics and Automation"}
@STRING{IEEE_J_AFFC       = "{IEEE} Transactions on Affective Computing"}
@STRING{IEEE_J_ASE        = "{IEEE} Transactions on Automation Science and Engineering"}
@STRING{IEEE_J_AMD        = "{IEEE} Transactions on Autonomous Mental Development"}
@STRING{IEEE_J_CCN        = "{IEEE} Transactions on Cognitive Communications and Networking"}
in 2015 AMD became CDS
@STRING{IEEE_J_CDS        = "{IEEE} Transactions on Cognitive and Developmental Systems"}
@STRING{IEEE_J_CIAIG      = "{IEEE} Transactions on Computational Intelligence and {AI} in Games"}
in 2013 SMCB became CYB
@STRING{IEEE_J_CYB        = "{IEEE} Transactions on Cybernetics"}
@STRING{IEEE_J_H          = "{IEEE} Transactions on Haptics"}
@STRING{IEEE_J_HFE        = "{IEEE} Transactions on Human Factors in Electronics"}
in 2013 SMCC became HMS
@STRING{IEEE_J_HMS        = "{IEEE} Transactions on Human-Machine Systems"}
@STRING{IEEE_J_MMS        = "{IEEE} Transactions on Man-Machine Systems"}
@STRING{IEEE_J_PAMI       = "{IEEE} Transactions on Pattern Analysis and Machine Intelligence"}
in 1989 JRA became RA
in August 2004, RA split into ASE and RO
@STRING{IEEE_J_RA         = "{IEEE} Transactions on Robotics and Automation"}
@STRING{IEEE_J_RAL        = "{IEEE} Robotics and Automation Letters"}
@STRING{IEEE_J_RO         = "{IEEE} Transactions on Robotics"}
@STRING{IEEE_J_SMC        = "{IEEE} Transactions on Systems, Man, and Cybernetics"}
@STRING{IEEE_J_SMCA       = "{IEEE} Transactions on Systems, Man, and Cybernetics---Part {A}: Systems and Humans"}
@STRING{IEEE_J_SMCB       = "{IEEE} Transactions on Systems, Man, and Cybernetics---Part {B}: Cybernetics"}
@STRING{IEEE_J_SMCC       = "{IEEE} Transactions on Systems, Man, and Cybernetics---Part {C}: Applications and Reviews"}
in 2012 SMCA became SMCS
@STRING{IEEE_J_SMCS       = "{IEEE} Transactions on Systems, Man, and Cybernetics: Systems"}
@STRING{IEEE_J_SSC        = "{IEEE} Transactions on Systems Science and Cybernetics"}



earth, wind, fire and water
@STRING{IEEE_J_GRSL       = "{IEEE} Geoscience and Remote Sensing Letters"}
@STRING{IEEE_J_GE         = "{IEEE} Transactions on Geoscience Electronics"}
@STRING{IEEE_J_GRS        = "{IEEE} Transactions on Geoscience and Remote Sensing"}
@STRING{IEEE_J_OE         = "{IEEE} Journal of Oceanic Engineering"}
@STRING{IEEE_J_STARS      = "{IEEE} Journal of Selected Topics in Applied Earth Observations and Remote Sensing"}



education, engineering, history, IEEE, professional
@STRING{IEEE_J_CJECE      = "Canadian Journal of Electrical and Computer Engineering"}
@STRING{IEEE_J_PROC       = "Proceedings of the {IEEE}"}
@STRING{IEEE_J_RITA       = "{IEEE} Revista Iberoamericana de Technolog{\'{i}}as del Aprendizaje"}
@STRING{IEEE_J_EDU        = "{IEEE} Transactions on Education"}
@STRING{IEEE_J_EM         = "{IEEE} Transactions on Engineering Management"}
@STRING{IEEE_J_EWS        = "{IEEE} Transactions on Engineering Writing and Speech"}
@STRING{IEEE_J_LT         = "{IEEE} Transactions on Learning Technologies"}
@STRING{IEEE_J_PC         = "{IEEE} Transactions on Professional Communication"}



electromagnetics, antennas, EMI, magnetics and microwave
@STRING{IEEE_J_AWPL       = "{IEEE} Antennas and Wireless Propagation Letters"}
@STRING{IEEE_J_MAGL       = "{IEEE} Magnetics Letters"}
@STRING{IEEE_J_MGWL       = "{IEEE} Microwave and Guided Wave Letters"}
@STRING{IEEE_J_MWCL       = "{IEEE} Microwave and Wireless Components Letters"}
@STRING{IEEE_J_RFIC       = "{IEEE} {RFIC} Journal"}
@STRING{IEEE_J_RFID       = "{IEEE} {RFID} Journal"}
@STRING{IEEE_J_AP         = "{IEEE} Transactions on Antennas and Propagation"}
@STRING{IEEE_J_EMC        = "{IEEE} Transactions on Electromagnetic Compatibility"}
@STRING{IEEE_J_MAG        = "{IEEE} Transactions on Magnetics"}
@STRING{IEEE_J_MTT        = "{IEEE} Transactions on Microwave Theory and Techniques"}
@STRING{IEEE_J_RFI        = "{IEEE} Transactions on Radio Frequency Interference"}
@STRING{IEEE_J_TTHZ       = "{IEEE} Transactions on Terahertz Science and Technology"}
@STRING{IEEE_J_TJMJ       = "{IEEE} Translation Journal on Magnetics in Japan"}



energy, power and conversion
@STRING{IEEE_J_PHOT       = "{IEEE} Journal of Photovoltaics"}
@STRING{IEEE_J_PEL        = "{IEEE} Power Electronics Letters"}
@STRING{IEEE_J_PETS       = "{IEEE} Power and Energy Technology Systems Journal"}
@STRING{IEEE_J_ESTPE      = "{IEEE} Transactions on Emerging and Selected Topics in Power Electronics"}
@STRING{IEEE_J_EC         = "{IEEE} Transactions on Energy Conversion"}
@STRING{IEEE_J_PWRAS      = "{IEEE} Transactions on Power Apparatus and Systems"}
@STRING{IEEE_J_PWRD       = "{IEEE} Transactions on Power Delivery"}
@STRING{IEEE_J_PWRE       = "{IEEE} Transactions on Power Electronics"}
@STRING{IEEE_J_PWRS       = "{IEEE} Transactions on Power Systems"}
@STRING{IEEE_J_SG         = "{IEEE} Transactions on Smart Grid"}
@STRING{IEEE_J_STE        = "{IEEE} Transactions on Sustainable Energy"}



industrial, commercial and consumer
@STRING{IEEE_J_PSE        = "{IEEE} Journal of Product Safety Engineering"}
@STRING{IEEE_J_APPIND     = "{IEEE} Transactions on Applications and Industry"}
@STRING{IEEE_J_BC         = "{IEEE} Transactions on Broadcasting"}
@STRING{IEEE_J_BCTV       = "{IEEE} Transactions on Broadcast and Television Receivers"}
@STRING{IEEE_J_CE         = "{IEEE} Transactions on Consumer Electronics"}
@STRING{IEEE_J_IE         = "{IEEE} Transactions on Industrial Electronics"}
@STRING{IEEE_J_IECI       = "{IEEE} Transactions on Industrial Electronics and Control Instrumentation"}
@STRING{IEEE_J_IA         = "{IEEE} Transactions on Industry Applications"}
@STRING{IEEE_J_IGA        = "{IEEE} Transactions on Industry and General Applications"}
@STRING{IEEE_J_IINF       = "{IEEE} Transactions on Industrial Informatics"}



instrumentation and measurement
@STRING{IEEE_J_IM         = "{IEEE} Transactions on Instrumentation and Measurement"}



insulation and materials
@STRING{IEEE_J_JEM        = "{IEEE/TMS} Journal of Electronic Materials"}
@STRING{IEEE_J_DEI        = "{IEEE} Transactions on Dielectrics and Electrical Insulation"}
@STRING{IEEE_J_EI         = "{IEEE} Transactions on Electrical Insulation"}



mechanical
@STRING{IEEE_J_MEMS       = "{IEEE/ASME} Journal of Microelectromechanical Systems"}
in 2014 MEMS became MEMSI
@STRING{IEEE_J_MEMSI      = "{IEEE} Journal of Microelectromechanical Systems"}
@STRING{IEEE_J_MECH       = "{IEEE/ASME} Transactions on Mechatronics"}



medical and biological
@STRING{IEEE_J_BHI        = "{IEEE} Journal of Biomedical and Health Informatics"}
@STRING{IEEE_J_TEHM       = "{IEEE} Journal of Translational Engineering in Health and Medicine"}
@STRING{IEEE_J_LS         = "{IEEE} Life Sciences Letters"}
@STRING{IEEE_J_RBME       = "{IEEE} Reviews in Biomedical Engineering"}
@STRING{IEEE_J_BCAS       = "{IEEE} Transactions on Biomedical Circuits and Systems"}
@STRING{IEEE_J_BMELC      = "{IEEE} Transactions on Bio-Medical Electronics"}
B-ME later dropped the hyphen and became the BME
@STRING{IEEE_J_BME        = "{IEEE} Transactions on Biomedical Engineering"}
@STRING{IEEE_J_B-ME       = "{IEEE} Transactions on Bio-Medical Engineering"}
@STRING{IEEE_J_CBB        = "{IEEE/ACM} Transactions on Computational Biology and Bioinformatics"}
@STRING{IEEE_J_ITBM       = "{IEEE} Transactions on Information Technology in Biomedicine"}
@STRING{IEEE_J_ME         = "{IEEE} Transactions on Medical Electronics"}
@STRING{IEEE_J_MI         = "{IEEE} Transactions on Medical Imaging"}
@STRING{IEEE_J_MBSC       = "{IEEE} Transactions on Molecular, Biological and Multi-Scale Communications"}
@STRING{IEEE_J_NB         = "{IEEE} Transactions on NanoBioscience"}
@STRING{IEEE_J_NSRE       = "{IEEE} Transactions on Neural Systems and Rehabilitation Engineering"}
@STRING{IEEE_J_RE         = "{IEEE} Transactions on Rehabilitation Engineering"}



optics, lightwave and photonics
@STRING{IEEE_J_JLT        = "{IEEE/OSA} Journal of Lightwave Technology"}
@STRING{IEEE_J_OCN        = "{IEEE} Journal of Optical Communications and Networking"}
@STRING{IEEE_J_PJ         = "{IEEE} Photonics Journal"}
@STRING{IEEE_J_PTL        = "{IEEE} Photonics Technology Letters"}



physics, electrons, nanotechnology, nuclear and quantum electronics
@STRING{IEEE_J_EDL        = "{IEEE} Electron Device Letters"}
@STRING{IEEE_J_EDS        = "{IEEE} Journal of Electron Devices Society"}
@STRING{IEEE_J_JQE        = "{IEEE} Journal of Quantum Electronics"}
@STRING{IEEE_J_JSTQE      = "{IEEE} Journal of Selected Topics in Quantum Electronics"}
@STRING{IEEE_J_ENANO      = "{IEEE} Nanotechnology Express"}
@STRING{IEEE_J_ED         = "{IEEE} Transactions on Electron Devices"}
@STRING{IEEE_J_NANO       = "{IEEE} Transactions on Nanotechnology"}
@STRING{IEEE_J_NS         = "{IEEE} Transactions on Nuclear Science"}
@STRING{IEEE_J_PS         = "{IEEE} Transactions on Plasma Science"}



reliability
@STRING{IEEE_J_DMR        = "{IEEE} Transactions on Device and Materials Reliability"}
@STRING{IEEE_J_R          = "{IEEE} Transactions on Reliability"}



semiconductors, superconductors, electrochemical and solid state
@STRING{IEEE_J_ESSL       = "{IEEE/ECS} Electrochemical and Solid-State Letters"}
@STRING{IEEE_J_JSSC       = "{IEEE} Journal of Solid-State Circuits"}
@STRING{IEEE_J_ASC        = "{IEEE} Transactions on Applied Superconductivity"}
@STRING{IEEE_J_SM         = "{IEEE} Transactions on Semiconductor Manufacturing"}



sensors
@STRING{IEEE_J_SENSOR     = "{IEEE} Sensors Journal"}



VLSI
@STRING{IEEE_J_VLSI       = "{IEEE} Transactions on Very Large Scale Integration ({VLSI}) Systems"}






IEEE Magazines and Online Publications


@STRING{IEEE_O_ACC        = "{IEEE} Access"}
@STRING{IEEE_M_AES        = "{IEEE} Aerospace and Electronics Systems Magazine"}
@STRING{IEEE_M_HIST       = "{IEEE} Annals of the History of Computing"}
@STRING{IEEE_M_AP         = "{IEEE} Antennas and Propagation Magazine"}
@STRING{IEEE_M_ASSP       = "{IEEE} {ASSP} Magazine"}
@STRING{IEEE_M_CHINAC     = "China Communications Magazine"}
@STRING{IEEE_M_CD         = "{IEEE} Circuits and Devices Magazine"}
@STRING{IEEE_M_CAS        = "{IEEE} Circuits and Systems Magazine"}
@STRING{IEEE_M_COM        = "{IEEE} Communications Magazine"}
@STRING{IEEE_M_COMSOC     = "{IEEE} Communications Society Magazine"}
@STRING{IEEE_O_CSTO       = "{IEEE} Communications Surveys and Tutorials"}
@STRING{IEEE_M_CIM        = "{IEEE} Computational Intelligence Magazine"}
CSEM changed to CSE in 1999
@STRING{IEEE_M_CSE        = "{IEEE} Computing in Science and Engineering"}
@STRING{IEEE_M_CSEM       = "{IEEE} Computational Science and Engineering Magazine"}
@STRING{IEEE_M_C          = "Computer"}
@STRING{IEEE_M_CAP        = "{IEEE} Computer Applications in Power"}
@STRING{IEEE_M_CGA        = "{IEEE} Computer Graphics and Applications"}
@STRING{IEEE_M_CONC       = "{IEEE} Concurrency"}
@STRING{IEEE_M_CS         = "{IEEE} Control Systems Magazine"}
in 2012 DTC became DT
@STRING{IEEE_M_DT         = "{IEEE} Design \&\ Test"}
@STRING{IEEE_M_DTC        = "{IEEE} Design and Test of Computers"}
@STRING{IEEE_O_DSO        = "{IEEE} Distributed Systems Online"}
@STRING{IEEE_M_EI         = "{IEEE} Electrical Insulation Magazine"}
@STRING{IEEE_M_ETF        = "{IEEE} Electrification Magazine"}
@STRING{IEEE_M_EMC        = "{IEEE} Electromagnetic Compatibility Magazine"}
@STRING{IEEE_M_ETR        = "{IEEE} ElectroTechnology Review"}
@STRING{IEEE_M_EMR        = "{IEEE} Engineering Management Review"}
@STRING{IEEE_M_EMB        = "{IEEE} Engineering in Medicine and Biology Magazine"}
@STRING{IEEE_M_EXP        = "{IEEE} Expert"}
@STRING{IEEE_M_GRS        = "{IEEE} Geoscience and Remote Sensing Magazine"}
@STRING{IEEE_M_IA         = "{IEEE} Industry Applications Magazine"}
@STRING{IEEE_M_IE         = "{IEEE} Industrial Electronics Magazine"}
@STRING{IEEE_M_IM         = "{IEEE} Instrumentation and Measurement Magazine"}
@STRING{IEEE_M_IS         = "{IEEE} Intelligent Systems"}
@STRING{IEEE_M_ITS        = "{IEEE} Intelligent Transportation Systems Magazine"}
@STRING{IEEE_M_IC         = "{IEEE} Internet Computing"}
@STRING{IEEE_M_ITP        = "{IEEE} {IT} Professional"}
@STRING{IEEE_M_MICRO      = "{IEEE} Micro"}
@STRING{IEEE_M_MW         = "{IEEE} Microwave Magazine"}
@STRING{IEEE_M_MM         = "{IEEE} Multimedia"}
@STRING{IEEE_M_NANO       = "{IEEE} Nanotechnology Magazine"}
@STRING{IEEE_M_NET        = "{IEEE} Network"}
@STRING{IEEE_M_PCOM       = "{IEEE} Personal Communications Magazine"}
@STRING{IEEE_M_PVC        = "{IEEE} Pervasive Computing"}
@STRING{IEEE_M_POT        = "{IEEE} Potentials"}
@STRING{IEEE_M_PEL        = "{IEEE} Power Electronics Magazine"}
CAP and PER merged to form PE in 2003
@STRING{IEEE_M_PE         = "{IEEE} Power and Energy Magazine"}
@STRING{IEEE_M_PER        = "{IEEE} Power Engineering Review"}
@STRING{IEEE_M_PULSE      = "{IEEE} Pulse"}
@STRING{IEEE_M_RA         = "{IEEE} Robotics and Automation Magazine"}
@STRING{IEEE_M_SAP        = "{IEEE} Security and Privacy"}
@STRING{IEEE_M_SP         = "{IEEE} Signal Processing Magazine"}
@STRING{IEEE_M_S          = "{IEEE} Software"}
@STRING{IEEE_M_SSC        = "{IEEE} Solid-State Circuits Magazine"}
@STRING{IEEE_M_SPECT      = "{IEEE} Spectrum"}
@STRING{IEEE_M_SMC        = "{IEEE} Systems, Man, and Cybernetics Magazine"}
@STRING{IEEE_M_TS         = "{IEEE} Technology and Society Magazine"}
@STRING{IEEE_M_VT         = "{IEEE} Vehicular Technology Magazine"}
@STRING{IEEE_M_WC         = "{IEEE} Wireless Communications"}
@STRING{IEEE_M_TODAY      = "Today's Engineer"}





--
EOF


\begin{thebibliography}{10}
\providecommand{\url}[1]{#1}
\csname url@samestyle\endcsname
\providecommand{\newblock}{\relax}
\providecommand{\bibinfo}[2]{#2}
\providecommand{\BIBentrySTDinterwordspacing}{\spaceskip=0pt\relax}
\providecommand{\BIBentryALTinterwordstretchfactor}{4}
\providecommand{\BIBentryALTinterwordspacing}{\spaceskip=\fontdimen2\font plus
\BIBentryALTinterwordstretchfactor\fontdimen3\font minus
  \fontdimen4\font\relax}
\providecommand{\BIBforeignlanguage}[2]{{%
\expandafter\ifx\csname l@#1\endcsname\relax
\typeout{** WARNING: IEEEtran.bst: No hyphenation pattern has been}%
\typeout{** loaded for the language `#1'. Using the pattern for}%
\typeout{** the default language instead.}%
\else
\language=\csname l@#1\endcsname
\fi
#2}}
\providecommand{\BIBdecl}{\relax}
\BIBdecl

\bibitem{kretzschmar2014learning}
H.~Kretzschmar, M.~Kuderer, and W.~Burgard, ``Learning to predict trajectories
  of cooperatively navigating agents,'' in \emph{Robotics and Automation
  (ICRA), 2014 IEEE International Conference on}.\hskip 1em plus 0.5em minus
  0.4em\relax IEEE, 2014, pp. 4015--4020.

\bibitem{koppula2016anticipating}
H.~S. Koppula and A.~Saxena, ``Anticipating human activities using object
  affordances for reactive robotic response,'' \emph{IEEE transactions on
  pattern analysis and machine intelligence}, vol.~38, no.~1, pp. 14--29, 2016.

\bibitem{bartoli2017context}
F.~Bartoli, G.~Lisanti, L.~Ballan, and A.~Del~Bimbo, ``Context-aware trajectory
  prediction,'' \emph{arXiv preprint arXiv:1705.02503}, 2017.

\bibitem{ellis2009modelling}
D.~Ellis, E.~Sommerlade, and I.~Reid, ``Modelling pedestrian trajectory
  patterns with gaussian processes,'' in \emph{Computer Vision Workshops (ICCV
  Workshops), 2009 IEEE 12th International Conference on}.\hskip 1em plus 0.5em
  minus 0.4em\relax IEEE, 2009, pp. 1229--1234.

\bibitem{kim2011gaussian}
K.~Kim, D.~Lee, and I.~Essa, ``Gaussian process regression flow for analysis of
  motion trajectories,'' in \emph{Computer vision (ICCV), 2011 IEEE
  international conference on}.\hskip 1em plus 0.5em minus 0.4em\relax IEEE,
  2011, pp. 1164--1171.

\bibitem{robicquet2016learning}
A.~Robicquet, A.~Sadeghian, A.~Alahi, and S.~Savarese, ``Learning social
  etiquette: Human trajectory understanding in crowded scenes,'' in
  \emph{European conference on computer vision}.\hskip 1em plus 0.5em minus
  0.4em\relax Springer, 2016, pp. 549--565.

\bibitem{vemula2018social}
A.~Vemula, K.~Muelling, and J.~Oh, ``Social attention: Modeling attention in
  human crowds,'' in \emph{Proceedings of the International Conference on
  Robotics and Automation (ICRA) 2018}, May 2018.

\bibitem{alahi2016social}
A.~Alahi, K.~Goel, V.~Ramanathan, A.~Robicquet, L.~Fei-Fei, and S.~Savarese,
  ``Social lstm: Human trajectory prediction in crowded spaces,'' in
  \emph{Proceedings of the IEEE Conference on Computer Vision and Pattern
  Recognition}, 2016, pp. 961--971.

\bibitem{gupta2018social}
A.~Gupta, J.~Johnson, L.~Fei-Fei, S.~Savarese, and A.~Alahi, ``Social gan:
  Socially acceptable trajectories with generative adversarial networks,'' in
  \emph{IEEE Conference on Computer Vision and Pattern Recognition (CVPR)}, no.
  CONF, 2018.

\bibitem{sadeghiankosaraju2018trajnet}
A.~Sadeghian, V.~Kosaraju, A.~Gupta, S.~Savarese, and A.~Alahi, ``Trajnet:
  Towards a benchmark for human trajectory prediction,'' \emph{arXiv preprint},
  2018.

\bibitem{jain2016structural}
A.~Jain, A.~R. Zamir, S.~Savarese, and A.~Saxena, ``Structural-rnn: Deep
  learning on spatio-temporal graphs,'' in \emph{Proceedings of the IEEE
  Conference on Computer Vision and Pattern Recognition}, 2016, pp. 5308--5317.

\bibitem{velickovic2017graph}
P.~Velickovic, G.~Cucurull, A.~Casanova, A.~Romero, P.~Lio, and Y.~Bengio,
  ``Graph attention networks,'' \emph{stat}, vol. 1050, p.~20, 2017.

\bibitem{vaswani2017attention}
A.~Vaswani, N.~Shazeer, N.~Parmar, J.~Uszkoreit, L.~Jones, A.~N. Gomez,
  {\L}.~Kaiser, and I.~Polosukhin, ``Attention is all you need,'' in
  \emph{Advances in Neural Information Processing Systems}, 2017, pp.
  6000--6010.

\bibitem{van2018relational}
S.~van Steenkiste, M.~Chang, K.~Greff, and J.~Schmidhuber, ``Relational neural
  expectation maximization: Unsupervised discovery of objects and their
  interactions,'' \emph{arXiv preprint arXiv:1802.10353}, 2018.

\bibitem{battaglia2016interaction}
P.~Battaglia, R.~Pascanu, M.~Lai, D.~J. Rezende \emph{et~al.}, ``Interaction
  networks for learning about objects, relations and physics,'' in
  \emph{Advances in neural information processing systems}, 2016, pp.
  4502--4510.

\bibitem{fernando2018soft+}
T.~Fernando, S.~Denman, S.~Sridharan, and C.~Fookes, ``Soft+ hardwired
  attention: An lstm framework for human trajectory prediction and abnormal
  event detection,'' \emph{Neural networks}, vol. 108, pp. 466--478, 2018.

\bibitem{helbing1995social}
D.~Helbing and P.~Molnar, ``Social force model for pedestrian dynamics,''
  \emph{Physical review E}, vol.~51, no.~5, p. 4282, 1995.

\bibitem{varshneya2017human}
D.~Varshneya and G.~Srinivasaraghavan, ``Human trajectory prediction using
  spatially aware deep attention models,'' \emph{arXiv preprint
  arXiv:1705.09436}, 2017.

\bibitem{lee2017desire}
N.~Lee, W.~Choi, P.~Vernaza, C.~B. Choy, P.~H. Torr, and M.~Chandraker,
  ``Desire: Distant future prediction in dynamic scenes with interacting
  agents,'' 2017.

\bibitem{scarselli2009graph}
F.~Scarselli, M.~Gori, A.~C. Tsoi, M.~Hagenbuchner, and G.~Monfardini, ``The
  graph neural network model,'' \emph{IEEE Transactions on Neural Networks},
  vol.~20, no.~1, pp. 61--80, 2009.

\bibitem{cho2014learning}
K.~Cho, B.~Van~Merri{\"e}nboer, C.~Gulcehre, D.~Bahdanau, F.~Bougares,
  H.~Schwenk, and Y.~Bengio, ``Learning phrase representations using rnn
  encoder-decoder for statistical machine translation,'' \emph{arXiv preprint
  arXiv:1406.1078}, 2014.

\bibitem{bahdanau2014neural}
D.~Bahdanau, K.~Cho, and Y.~Bengio, ``Neural machine translation by jointly
  learning to align and translate,'' \emph{arXiv preprint arXiv:1409.0473},
  2014.

\bibitem{song2017end}
S.~Song, C.~Lan, J.~Xing, W.~Zeng, and J.~Liu, ``An end-to-end spatio-temporal
  attention model for human action recognition from skeleton data.'' in
  \emph{AAAI}, vol.~1, no.~2, 2017, p.~7.

\bibitem{liu2016spatio}
J.~Liu, A.~Shahroudy, D.~Xu, and G.~Wang, ``Spatio-temporal lstm with trust
  gates for 3d human action recognition,'' in \emph{European Conference on
  Computer Vision}.\hskip 1em plus 0.5em minus 0.4em\relax Springer, 2016, pp.
  816--833.

\bibitem{xue2018ss}
H.~Xue, D.~Q. Huynh, and M.~Reynolds, ``Ss-lstm: A hierarchical lstm model for
  pedestrian trajectory prediction,'' in \emph{2018 IEEE Winter Conference on
  Applications of Computer Vision (WACV)}.\hskip 1em plus 0.5em minus
  0.4em\relax IEEE, 2018, pp. 1186--1194.

\bibitem{liang2016semantic}
X.~Liang, X.~Shen, J.~Feng, L.~Lin, and S.~Yan, ``Semantic object parsing with
  graph lstm,'' in \emph{European Conference on Computer Vision}.\hskip 1em
  plus 0.5em minus 0.4em\relax Springer, 2016, pp. 125--143.

\bibitem{yuan2017temporal}
Y.~Yuan, X.~Liang, X.~Wang, D.~Y. Yeung, and A.~Gupta, ``Temporal dynamic graph
  lstm for action-driven video object detection,'' \emph{arXiv preprint
  arXiv:1708.00666}, 2017.

\bibitem{pellegrini2009you}
S.~Pellegrini, A.~Ess, K.~Schindler, and L.~Van~Gool, ``You'll never walk
  alone: Modeling social behavior for multi-target tracking,'' in
  \emph{Computer Vision, 2009 IEEE 12th International Conference on}.\hskip 1em
  plus 0.5em minus 0.4em\relax IEEE, 2009, pp. 261--268.

\bibitem{lerner2007crowds}
A.~Lerner, Y.~Chrysanthou, and D.~Lischinski, ``Crowds by example,'' in
  \emph{Computer Graphics Forum}, vol.~26, no.~3.\hskip 1em plus 0.5em minus
  0.4em\relax Wiley Online Library, 2007, pp. 655--664.

\end{thebibliography}



\markboth{Journal of \LaTeX\ Class Files,~Vol.~6, No.~1, January~2007}%
{Shell \MakeLowercase{\textit{et al.}}: Bare Demo of IEEEtran.cls for Journals}

\end{document}